\documentclass[lettersize,journal]{IEEEtran}
\usepackage[utf8]{inputenc}
\usepackage{amssymb}
\usepackage{amsmath,amsfonts}
\usepackage{bbm}
\usepackage{algorithmic}
\usepackage{array}
\usepackage[caption=false,font=normalsize,labelfont=sf,textfont=sf]{subfig}
\usepackage{textcomp}
\usepackage{stfloats}
\usepackage{url}
\usepackage{xcolor}
\usepackage{verbatim}
\usepackage{graphicx}
\usepackage{multirow} 
\usepackage{booktabs}
\def\BibTeX{{\rm B\kern-.05em{\sc i\kern-.025em b}\kern-.08em
    T\kern-.1667em\lower.7ex\hbox{E}\kern-.125emX}}
\usepackage{balance}
\begin{document}
\title{FORTRESS: Function-composition Optimized Real-Time Resilient Structural Segmentation via Kolmogorov-Arnold Enhanced Spatial Attention Networks}
\author{Christina Thrainer,~\IEEEmembership{}
        Md Meftahul Ferdaus,~\IEEEmembership{}
        Mahdi Abdelguerfi,~\IEEEmembership{}
        Christian Guetl,~\IEEEmembership{}
        Steven Sloan,~\IEEEmembership{}
        Kendall N. Niles,~\IEEEmembership{}and~Ken Pathak~\IEEEmembership{}
        \thanks{M. Ferdaus and M. Abdelguerfi are with the Canizaro Livingston Gulf States Center for Environmental Informatics, the University of New Orleans, New Orleans, USA (e-mail: mferdaus@uno.edu; gulfsceidirector@uno.edu).}
        \thanks{C. Thrainer and C. Guetl are with the Graz University of Technology, Graz, Austria.}
        \thanks{S. Sloan, K. N. Niles, and K. Pathak are with the US Army Corps of Engineers, Engineer Research and Development Center, Vicksburg, Mississippi, USA.}
        \thanks{Manuscript received July XX, 2025; revised July XX, 2025.}}

\markboth{Journal of \LaTeX\ Class Files,~Vol.~18, No.~9, September~2020}%
{How to Use the IEEEtran \LaTeX \ Templates}

\maketitle

\begin{abstract}
Automated structural defect segmentation in civil infrastructure faces a critical challenge: achieving high accuracy while maintaining computational efficiency for real-time deployment. This paper presents FORTRESS (Function-composition Optimized Real-Time Resilient Structural Segmentation), a new architecture that balances accuracy and speed by using a special method that combines depthwise separable convolutions with adaptive Kolmogorov-Arnold Network integration. FORTRESS incorporates three key innovations: a systematic depthwise separable convolution framework achieving a 3.6× parameter reduction per layer, adaptive TiKAN integration that selectively applies function composition transformations only when computationally beneficial, and multi-scale attention fusion combining spatial, channel, and KAN-enhanced features across decoder levels. The architecture achieves remarkable efficiency gains with 91\% parameter reduction (31M → 2.9M), 91\% computational complexity reduction (13.7 → 1.17 GFLOPs), and 3× inference speed improvement while delivering superior segmentation performance. Evaluation on benchmark infrastructure datasets demonstrates state-of-the-art results with an F1-score of 0.771 and a mean IoU of 0.677, significantly outperforming existing methods including U-Net, SA-UNet, and U-KAN. The dual optimization strategy proves essential for optimal performance, establishing FORTRESS as a robust solution for practical structural defect segmentation in resource-constrained environments where both accuracy and computational efficiency are paramount. Comprehensive architectural specifications are provided in the Supplemental Material. Source code is available at URL: \url{https://github.com/faeyelab/fortress-paper-code}.
\end{abstract}


\section{Introduction}
\IEEEPARstart{T}{he} deterioration of civil infrastructure represents one of the most pressing challenges facing modern society, with aging bridges, tunnels, buildings, and transportation networks requiring continuous monitoring to ensure public safety and operational efficiency. Traditional manual inspection methods, while thorough, are inherently time-consuming, labor-intensive, and subject to human error, making them inadequate for the scale and frequency of monitoring required by contemporary infrastructure systems \cite{jha2023deep, xu2024few}. The emergence of automated inspection technologies, particularly those leveraging computer vision and deep learning, has opened new possibilities for comprehensive, objective, and cost-effective structural health monitoring \cite{zhou2024defect, alexander2022fusion}.

Structural defects in civil infrastructure manifest in various forms, including cracks, spalling, corrosion, deformation, and material degradation, each presenting unique challenges for automated detection and segmentation systems. These defects often exhibit complex geometric patterns, variable scales, and subtle visual characteristics that can be easily obscured by environmental factors such as lighting conditions, surface textures, and debris accumulation \cite{alexander2022fusion, midwinter2025learning}. Furthermore, the critical nature of infrastructure safety demands that automated inspection systems achieve not only high accuracy but also reliable real-time performance to enable timely maintenance interventions and prevent catastrophic failures \cite{xu2024few, zhou2024defect}.

Recent advances in deep learning, particularly in semantic segmentation, have demonstrated significant potential for addressing these challenges \cite{arafin2023deep, dang2023lightweight}. Fully Convolutional Networks (FCNs) and U-Net-based architectures have emerged as dominant approaches for pixel-level defect localization, offering the spatial precision necessary for detailed structural analysis \cite{alkayem2024cocracksegment, sabet2022automated,panta2023iterlunet}. However, these conventional approaches face fundamental limitations when applied to infrastructure inspection scenarios. The complexity of structural defect patterns often requires extensive computational resources, making deployment in field conditions challenging. Additionally, the class imbalance inherent in defect detection tasks, where defective regions typically constitute a small fraction of the total image area, poses significant challenges for training robust and generalizable models \cite{mahmoudi2025addressing}.

The recent introduction of Kolmogorov-Arnold Networks (KANs) has opened new avenues for addressing these limitations through function composition-based learning paradigms \cite{moradi2024kan, wu2025mofkan,ferdaus2024kanice}. Unlike traditional neural networks that rely primarily on matrix multiplications and fixed activation functions, KANs model complex relationships through compositions of learnable one-dimensional functions, potentially offering more efficient and interpretable representations for complex pattern recognition tasks \cite{seydi2025hyperspectral, zhou2024whitebox}. The theoretical foundation of KANs, rooted in the Kolmogorov-Arnold representation theorem, suggests that any continuous multivariate function can be represented as a composition of continuous univariate functions, providing a mathematically principled approach to function approximation that may be particularly well-suited for the complex, multi-scale nature of structural defect patterns \cite{brateanu2024lowlight}.

Building upon these insights, this paper introduces FORTRESS (Function-composition Optimized Real-Time Resilient Structural Segmentation), a novel hybrid architecture that synergistically combines the spatial attention mechanisms of SA-UNet with the function composition capabilities of TiKAN (parameter-efficient Kolmogorov-Arnold Networks) to achieve superior performance in structural defect segmentation tasks. The proposed approach addresses the fundamental trade-off between computational efficiency and segmentation accuracy that has limited the practical deployment of deep learning-based inspection systems in real-world infrastructure monitoring scenarios.

The key contributions of this work are threefold. First, we introduce a novel dual optimization strategy that systematically combines depthwise separable convolutions with adaptive Kolmogorov-Arnold Network (KAN) integration, establishing a new paradigm for efficient structural defect segmentation that achieves both superior accuracy and remarkable computational efficiency. This approach addresses the fundamental trade-off between segmentation performance and computational overhead that has limited practical deployment of deep learning-based infrastructure inspection systems. Second, we present a comprehensive parameter efficiency framework that achieves 91\% parameter reduction (from 31M to 2.9M parameters) and 3× inference speed improvement while preserving representational capacity through selective function composition transformations applied only when computationally beneficial (spatial resolution $\leq$ 1024 pixels). Third, we demonstrate through extensive experimental evaluation on benchmark infrastructure datasets that FORTRESS achieves state-of-the-art segmentation performance (F1-score: 0.771, mIoU: 0.677) while requiring significantly fewer computational resources than existing methods, validating the effectiveness of the dual optimization paradigm for practical real-time deployment in resource-constrained inspection environments. Detailed architectural analysis and implementation specifications provided in the supplementary material to ensure reproducibility and facilitate future research.

The remainder of this paper is organized as follows. Section~\ref{sec:related_work} reviews related work on defect segmentation and relevant network architectures. Section~\ref{sec:prob_statement} formulates the research problem. Section~\ref{sec:methodology} details the proposed FORTRESS architecture. Section~\ref{sec:data-n-exp} describes the datasets, experimental setup, and evaluation protocols. Section~\ref{sec:results-n-discussion} presents and discusses the experimental results, including comparative analyses and ablation studies. Finally, Section~\ref{sec:conclusion} concludes the paper and outlines future research directions.

\section{Related Work}\label{sec:related_work}

\subsection{Structural Defect Detection and Segmentation}

The field of automated structural defect detection has evolved significantly over the past decade, driven by advances in computer vision and the increasing availability of high-resolution imaging technologies \cite{bhattacharya2022stand}. Early approaches relied primarily on traditional image processing techniques, including edge detection, morphological operations, and texture analysis, to identify potential defect regions. While these methods provided valuable insights into the fundamental characteristics of structural defects, they were limited by their reliance on hand-crafted features and their inability to adapt to the diverse range of defect manifestations encountered in real-world infrastructure.

The introduction of machine learning techniques marked a significant advancement in the field, with support vector machines, random forests, and other classical algorithms demonstrating improved performance over traditional image processing approaches \cite{xiao2024vibration, gunes2024localizing}. However, these methods still required extensive feature engineering and domain expertise to achieve satisfactory results, limiting their scalability and generalizability across different types of infrastructure and environmental conditions \cite{rajpoot2025predictive, samudra2023machine}.

The emergence of deep learning has revolutionized structural defect detection, with convolutional neural networks (CNNs) demonstrating unprecedented performance in both classification and segmentation tasks \cite{arafin2023deep, oulahyane2024assessing}. FCNs introduced by Long et al. \cite{long2015fully} established the foundation for pixel-level defect localization, enabling precise boundary delineation that is essential for quantitative structural analysis \cite{jin2025comparison, zhao2023development}. The U-Net architecture, originally developed for biomedical image segmentation, has been widely adopted for infrastructure inspection due to its ability to combine low-level detail preservation with high-level semantic understanding through skip connections \cite{arafin2023deep, jin2025comparison,alshawi2023dual}.

Recent work in this domain has focused on addressing specific challenges associated with the detection of structural defects. The KARMA architecture introduced parameter-efficient approaches for defect segmentation through Kolmogorov-Arnold representation learning, demonstrating significant improvements in computational efficiency while maintaining competitive accuracy. However, KARMA's focus on pure function composition approaches may limit its ability to capture the spatial relationships that are crucial for structural defect analysis.

Several researchers have explored attention mechanisms for improving defect detection performance \cite{ji2023textile, sui2023pddd}. Certain researchers have developed attention-guided networks \cite{zhang2018progressive} to concentrate on defects-relevant regions, whereas others have proposed dual-attention mechanisms that merge spatial and channel attention to improve feature representation \cite{liu2024wgsyolo, jiang2024adafnet}. These approaches have shown consistent improvements over baseline architectures, highlighting the importance of attention mechanisms to handle the complex and variable nature of structural defects \cite{sui2023pddd, ji2023textile}.

\subsection{Kolmogorov-Arnold Networks and Function Composition}

The theoretical foundation of Kolmogorov-Arnold Networks originates from the Kolmogorov-Arnold representation theorem, which states that any continuous multivariate function can be represented as a finite composition of continuous univariate functions \cite{basina2024kat, somvanshi2024survey}. This mathematical principle provides a fundamentally different approach to function approximation compared to traditional neural networks, which rely primarily on linear transformations followed by fixed activation functions \cite{liu2024kan}.

The practical implementation of KANs has been facilitated by recent advances in differentiable programming and automatic differentiation frameworks. Liu et al. \cite{liu2024kan} introduced the first practical KAN implementations, demonstrating their effectiveness on various function approximation tasks. The key insight underlying KANs is that complex multivariate relationships can be decomposed into simpler univariate functions, potentially leading to more interpretable and efficient representations.

We created TiKAN (Tiny Kolmogorov-Arnold Networks) as an important step forward in rendering KANs viable for extensive-scale implementations. By incorporating low-rank factorization and other parameter reduction techniques, TiKAN maintains the representational power of full KANs while dramatically reducing computational requirements. This efficiency gain is particularly important for computer vision applications, where the high dimensionality of image data can make standard KAN implementations computationally prohibitive.

Recent applications of KANs in computer vision have shown promising results across various domains \cite{cang2024kanvision}. Medical image analysis has been a particularly active area, with researchers demonstrating that KAN-based approaches can achieve competitive or superior performance compared to traditional CNNs while requiring significantly fewer parameters \cite{wang2025medkaformer, zeng2024mffkan}. However, the application of KANs to structural defect detection remains largely unexplored, representing a significant opportunity for advancing the state-of-the-art in infrastructure inspection \cite{krzywda2025metal}.

\subsection{Spatial Attention Mechanisms in Medical and Infrastructure Imaging}
Spatial attention mechanisms have emerged as a powerful tool for improving the performance of deep learning models in various computer vision tasks. The fundamental principle underlying spatial attention is the selective emphasis of relevant spatial regions while suppressing irrelevant or noisy areas, mimicking the human visual attention system's ability to focus on salient features.

The SA-UNet architecture represents \cite{guo2021sa} a significant advancement in applying spatial attention to medical image segmentation. By incorporating spatial attention modules at multiple scales within the U-Net framework, SA-UNet achieves improved performance in tasks requiring precise boundary delineation and fine-detail preservation. The success of SA-UNet in medical imaging suggests its potential applicability to structural defect detection, where similar challenges of complex boundary identification and multi-scale feature integration are encountered.

Channel attention mechanisms, such as those implemented in Squeeze-and-Excitation networks, provide complementary benefits by adaptively recalibrating channel-wise feature responses. The combination of spatial and channel attention has been shown to be particularly effective for tasks involving complex spatial relationships and multi-scale feature integration, making it well-suited for structural defect analysis \cite{chen2023unet,xiao2023gaei}.

\subsection{Bridging the Gap: Our Contribution}
Although the reviewed literature demonstrates significant progress, a clear gap remains in achieving both high segmentation accuracy and real-time computational efficiency for practical infrastructure inspection. Existing methods often excel in one area at the expense of the other. For example, traditional U-Net architectures and their attention-enhanced variants \cite{ronneberger2015u, guo2021sa} can achieve high accuracy but suffer from large parameter counts and high computational loads, making them unsuitable for resource-constrained deployment. In contrast, while some lightweight models \cite{ruan2023ege} prioritize efficiency, they often do so with a notable sacrifice in segmentation performance. Novel architectures like KANs are still in early application stages in this domain. \cite{li2024ukan}, and existing implementations have not yet fully optimized the balance between their powerful function-composition capabilities and the computational demands of high-resolution image segmentation. FORTRESS tackles this trade-off effectively. FORTRESS utilizes a dual optimization strategy, combining parameter-efficient separable convolutions with adaptive TiKAN integration, achieving state-of-the-art accuracy in a lightweight, fast architecture for real-time deployment.

\section{Problem Formulation}\label{sec:prob_statement}

Let $\mathcal{S} = \{S_i\}_{i=1}^N$ be a dataset of $N$ structural inspection images, where each image $S_i \in \mathbb{R}^{H \times W \times C}$ has a corresponding ground-truth segmentation mask $D_i \in \{0, 1\}^{H \times W \times K}$. Here, $H, W, C$ are the height, width, and channels of the image, and $K$ is the number of defect categories. The objective is to learn a mapping function $g_\phi: \mathcal{S} \rightarrow \mathcal{D}$, parameterized by $\phi$, that accurately segments diverse structural defects while maintaining computational efficiency for real-time deployment.

FORTRESS addresses this by minimizing a dual-optimization objective that balances segmentation accuracy with computational complexity:
\begin{equation}
\phi^* = \arg\min_\phi \left[ \mathcal{L}_{seg}(\phi) + \lambda_{eff} \mathcal{C}_{comp}(\phi) \right]
\end{equation}
where $\mathcal{L}_{seg}$ is the segmentation loss, $\mathcal{C}_{comp}$ is a penalty for computational complexity, and $\lambda_{eff}$ is a balancing hyperparameter. This strategy is realized through two core architectural principles: (1) systematic use of depthwise separable convolutions for parameter efficiency, and (2) adaptive integration of Kolmogorov-Arnold Networks (KANs) for enhanced feature representation.

The encoder pathway extracts hierarchical features $\{E_j\}_{j=1}^J$ at $J$ levels. The foundation of our efficiency-focused design is the depthwise separable convolution, which factors a standard convolution into two smaller operations. For an input feature map $E_{j-1}$, the output $E_j^{conv}$ is computed as:
\begin{equation}
E_j^{conv} = \mathcal{PW}_{1 \times 1}(\mathcal{DW}_{3 \times 3}(E_{j-1}; W_{dw}^{(j)}); W_{pw}^{(j)})
\end{equation}
where $\mathcal{DW}_{3 \times 3}$ is a depthwise convolution and $\mathcal{PW}_{1 \times 1}$ is a pointwise convolution. This factorization yields a significant parameter reduction, which for a $3 \times 3$ kernel is:
\begin{equation}
\text{Reduction Factor} = \frac{9 \cdot C_{in} \cdot C_{out}}{C_{in} \cdot (9 + C_{out})} \approx \frac{9 \cdot C_{out}}{9 + C_{out}}
\end{equation}


Within this efficient framework, we integrate multi-scale attention mechanisms. First, a channel attention module, $\mathcal{A}_{channel}$, recalibrates channel-wise feature responses:
\begin{equation}
\mathcal{A}_{channel}(E_j) = \sigma_{sigmoid}(\mathcal{MLP}(\text{GAP}(E_j)) + \mathcal{MLP}(\text{GMP}(E_j))) \odot E_j
\end{equation}
Second, a spatial attention module, $\mathcal{A}_{spatial}$, selectively emphasizes salient spatial regions. This module is also implemented efficiently using a depthwise separable convolution ($\mathcal{DS}_{k_d \times k_d}$):
\begin{align}
\tilde{E}_j &= \mathcal{A}_{spatial}(E_j) \nonumber \\
&= \sigma_{sigmoid}(\mathcal{DS}_{k_d \times k_d}([\text{Mean}(E_j); \text{Max}(E_j)])) \odot E_j
\end{align}

where $\tilde{E}_j$ represents the attention-refined feature map, $k_d \in \{3, 5, 7\}$ is the kernel size at decoder level $d$, and $\odot$ denotes element-wise multiplication.

The attention-refined features $\tilde{E}_j$ are then conditionally enhanced by an adaptive TiKAN module, $\tau_j$. This module applies powerful function-composition transformations only when computationally beneficial:
\begin{equation}
\hat{E}_j = \begin{cases}
\tilde{E}_j + \alpha_j \cdot \tau_j(\tilde{E}_j; \Phi_j) & \text{if } C_j \geq 16 \text{ and } H_j \times W_j \leq 1024 \\
\tilde{E}_j & \text{otherwise}
\end{cases}
\end{equation}
where $\hat{E}_j$ is the final enhanced feature map and $\alpha_j$ is a learnable scaling factor. The function $\tau_j$ models complex patterns using learnable spline-based univariate functions, defined as:
\begin{align}
\tau_j(x; \Phi_j) &= \sum_{k=1}^{K_j} w_{j,k} \cdot \nonumber \\
&\quad [\phi_{base}(x) \cdot s_{base} + \phi_{spline}(\mathcal{DW}_{enhance}(x)) \cdot s_{spline}]
\end{align}

The decoder pathway incorporates multi-scale fusion combining spatial attention, channel attention, and TiKAN-enhanced features:

\begin{equation}
R_d = \mathcal{F}_{decode}^{(d)}(\mathcal{A}_{spatial}^{(d)}(\mathcal{A}_{channel}(\hat{E}_{J-d+1})) \oplus \mathcal{U}(R_{d-1}))
\end{equation}

where $\mathcal{F}_{decode}^{(d)}$ represents decoder blocks with depthwise separable convolutions, $\oplus$ denotes concatenation, and $\mathcal{U}$ is upsampling. The complete architecture with deep supervision is formulated as:

\begin{equation}
\hat{D} = g_\phi(S) = \sigma(\mathcal{H}_{final}(R_1) + \sum_{d=2}^{4} \beta_d \cdot \mathcal{H}_d(R_d))
\end{equation}

where $\mathcal{H}_{final}$ and $\mathcal{H}_d$ are prediction heads, and $\beta_d \in \{0.4, 0.3, 0.2\}$ are deep supervision weights.

The training objective incorporates multi-scale supervision with class-specific weighting to handle dataset imbalance:

\begin{equation}
\mathcal{L}_{total} = \mathcal{L}_{CE}(\hat{D}, D) + \sum_{d=2}^{4} \beta_d \cdot \mathcal{L}_{CE}(\mathcal{H}_d(R_d), D_d)
\end{equation}

where the cross-entropy loss includes adaptive class weights $w_k$ computed based on inverse class frequency to handle dataset-specific class imbalance:

\begin{equation}
w_k = \frac{N_{total}}{\sum_{j=1}^{K} N_j} \cdot \frac{1}{N_k}
\end{equation}

where $N_k$ represents the number of pixels belonging to class $k$. Through this dual optimization strategy, FORTRESS achieves significant efficiency improvements while maintaining superior segmentation accuracy across diverse infrastructure datasets.

\section{Methodology}\label{sec:methodology}

\subsection{FORTRESS Architecture Overview}

\begin{figure}[!t]
\centering
\includegraphics[width=0.49\textwidth]{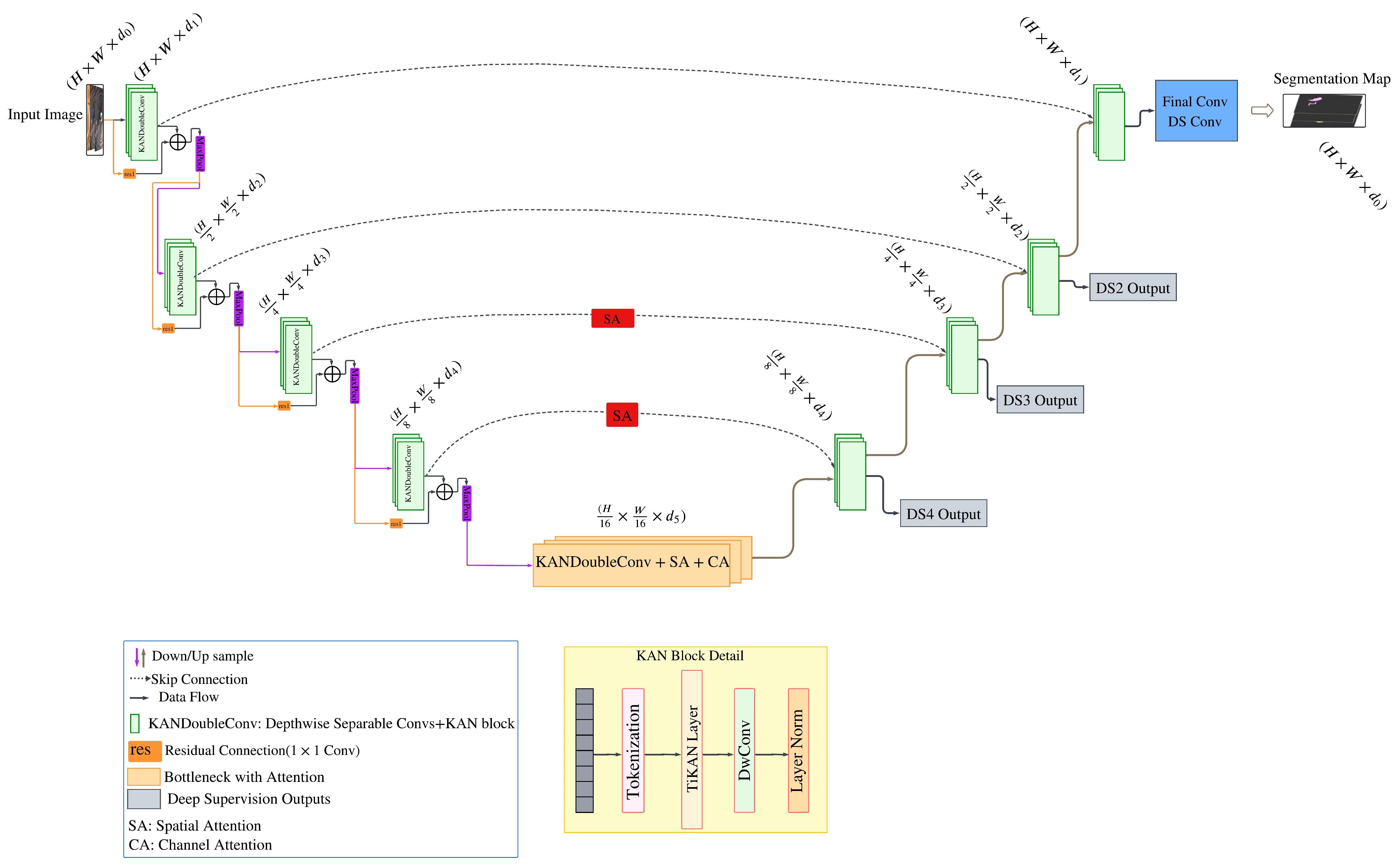}
\caption{FORTRESS architecture overview. The model employs a U-Net-inspired encoder-decoder structure with five levels of feature extraction and corresponding decoder pathways. Key innovations include: (1) systematic replacement of standard convolutions with depthwise separable convolutions for parameter efficiency, (2) adaptive TiKAN integration that selectively applies Kolmogorov-Arnold transformations based on feature characteristics, (3) multi-scale attention fusion combining spatial attention, channel attention, and KAN-enhanced features, and (4) deep supervision at multiple decoder levels for robust training. The dual optimization strategy achieves significant efficiency gains while maintaining superior segmentation accuracy.}
\label{fig:fortress_architecture}
\end{figure}

The FORTRESS architecture, illustrated in Figure~\ref{fig:fortress_architecture}, represents a novel synthesis of spatial attention mechanisms, Kolmogorov-Arnold representation learning, and depthwise separable convolutions, specifically designed to address the unique challenges of structural defect segmentation while maintaining computational efficiency suitable for real-time deployment. The overall architecture follows a U-Net-inspired encoder-decoder structure, enhanced with adaptive TiKAN integration, depthwise separable convolution optimization, and multi-scale attention fusion mechanisms that enable efficient and accurate defect localization across diverse infrastructure scenarios.

The architecture achieves significant parameter reduction through a dual optimization strategy. The first element of this strategy is the systematic replacement of standard convolutions with depthwise separable convolutions. A standard $3 \times 3$ convolution has a parameter cost of $C_{in} \times C_{out} \times 3 \times 3$. In contrast, a depthwise separable convolution has a cost of $(C_{in} \times 3 \times 3) + (C_{in} \times C_{out} \times 1 \times 1)$. 

The reduction factor is therefore:
\begin{equation}
\text{Reduction Factor} = \frac{C_{in} \times C_{out} \times 9}{C_{in} \times 9 + C_{in} \times C_{out}} = \frac{9 \cdot C_{out}}{9 + C_{out}}
\end{equation}
For a typical layer in our network where the number of output channels ($C_{out}$) is large (e.g., 64 or more), this factor approaches $\frac{9 \cdot C_{out}}{C_{out}} \approx 9$. In practice, across the entire architecture, this leads to a parameter reduction of approximately 3.6× for the convolutional layers.

The second element of the strategy is the adaptive TiKAN integration, which selectively applies function composition transformations only where computationally beneficial. This combination results in approximately 65\% parameter reduction compared to conventional attention-based U-Net variants while maintaining superior segmentation performance. While this section provides an overview, comprehensive technical details are provided in the supplementary material.

Given an input structural image $S \in \mathbb{R}^{H \times W \times 3}$, the FORTRESS architecture produces a segmentation prediction $\hat{D} \in \mathbb{R}^{H \times W \times K}$ through a hierarchical feature processing pipeline. The complete architecture can be mathematically expressed as:

\begin{equation}
\hat{D} = \mathcal{G}_{FORTRESS}(S; \Theta) = \mathcal{D}_{decoder}(\mathcal{E}_{encoder}(S; \Theta_E); \Theta_D)
\end{equation}

where $\mathcal{E}_{encoder}$ and $\mathcal{D}_{decoder}$ represent the encoder and decoder pathways respectively, and $\Theta = \{\Theta_E, \Theta_D\}$ denotes the complete set of learnable parameters optimized through depthwise separable convolution factorization.

The encoder pathway consists of five levels of feature extraction, each incorporating KANDoubleConv blocks that combine depthwise separable convolutional operations with adaptive TiKAN transformations:

\begin{equation}
E_j = \mathcal{F}_{KANDouble}^{(j)}(E_{j-1}; \Theta_j) + \mathcal{R}_j(E_{j-1}), \quad j = 1, 2, 3, 4, 5
\end{equation}

where $E_0 = S$, $\mathcal{F}_{KANDouble}^{(j)}$ represents the KANDoubleConv block at level $j$, and $\mathcal{R}_j$ denotes residual connections that facilitate gradient flow and feature preservation across the deep architecture.

\subsection{Enhanced KANDoubleConv Block Design with Depthwise Separable Convolutions}

The KANDoubleConv block represents the fundamental building unit of the FORTRESS architecture, designed to seamlessly integrate parameter-efficient depthwise separable convolutional operations with Kolmogorov-Arnold transformations. Each block processes input features through a sequence of operations that preserve spatial locality while enhancing representational capacity and achieving significant computational efficiency gains.

The core innovation of FORTRESS lies in the systematic replacement of standard convolutions with depthwise separable convolutions throughout the architecture. For an input feature map $F_{in} \in \mathbb{R}^{H \times W \times C_{in}}$, the depthwise separable convolutional processing is defined as:

\begin{equation}
F_{dw1} = \mathcal{DW}_{3 \times 3}(F_{in}; W_{dw1})
\end{equation}

\begin{equation}
F_{conv1} = \sigma_{ReLU}(\mathcal{BN}(\mathcal{PW}_{1 \times 1}(F_{dw1}; W_{pw1}, b_1)))
\end{equation}

\begin{equation}
F_{dw2} = \mathcal{DW}_{3 \times 3}(F_{conv1}; W_{dw2})
\end{equation}

\begin{equation}
F_{conv2} = \sigma_{ReLU}(\mathcal{BN}(\mathcal{PW}_{1 \times 1}(F_{dw2}; W_{pw2}, b_2)))
\end{equation}

where $\mathcal{DW}_{3 \times 3}$ denotes depthwise convolution with $3 \times 3$ kernels and $\mathcal{PW}_{1 \times 1}$ represents pointwise convolution with $1 \times 1$ kernels.

The TiKAN enhancement module operates conditionally based on feature characteristics and computational constraints. The adaptive activation criterion is defined as:

\begin{equation}
\text{TiKAN\_Active} = \begin{cases}
\text{True} & \text{if } C_{in} \geq 16 \text{ and } H \times W \leq 1024 \\
\text{False} & \text{otherwise}
\end{cases}
\end{equation}

When TiKAN is active, the feature transformation proceeds through patch embedding, KAN processing, and spatial reconstruction. The core TiKAN transformation follows the Kolmogorov-Arnold representation principle through learnable univariate functions enhanced with depthwise convolution operations.

\subsection{Enhanced Spatial Attention Mechanism}

The spatial attention mechanism in FORTRESS selectively emphasizes defect-relevant spatial regions while suppressing background noise, implemented using depthwise separable convolutions for computational efficiency. For an input feature map $F \in \mathbb{R}^{H \times W \times C}$, the spatial attention computation proceeds through parallel pooling operations and depthwise separable convolutional processing.

The spatial attention map is generated through:

\begin{equation}
F_{avg} = \text{Mean}(F, \text{dim}=[H, W]) \in \mathbb{R}^{1 \times 1 \times C}
\end{equation}

\begin{equation}
F_{max} = \text{Max}(F, \text{dim}=[H, W]) \in \mathbb{R}^{1 \times 1 \times C}
\end{equation}

The pooled features are concatenated and processed through a depthwise separable convolution:

\begin{equation}
F_{concat} = [F_{avg}; F_{max}] \in \mathbb{R}^{H \times W \times 2}
\end{equation}

\begin{equation}
A_{spatial} = \sigma_{sigmoid}(\mathcal{DS}_{k \times k}(F_{concat}; W_{ds}))
\end{equation}

FORTRESS employs multi-scale spatial attention with different kernel sizes at different decoder levels, enabling the capture of defect patterns at different spatial scales while maintaining computational efficiency.

\subsection{Enhanced Channel Attention Mechanism}

Complementing spatial attention, the channel attention mechanism adaptively recalibrates channel-wise feature responses based on their relevance to structural defect detection. For an input feature map $F \in \mathbb{R}^{H \times W \times C}$, channel statistics are computed through global pooling operations, and channel attention weights are generated through a shared multi-layer perceptron with a bottleneck structure optimized for the feature dimensions typical in depthwise separable convolution architectures.

\subsection{Multi-Scale Attention Fusion Framework}

The multi-scale attention fusion framework combines spatial attention, channel attention, and TiKAN-enhanced features across multiple decoder levels to achieve comprehensive defect pattern recognition while maintaining the computational benefits of depthwise separable convolutions. At each decoder level $d$, the multi-scale attention fusion is formulated as:

\begin{equation}
F_{fused}^{(d)} = \mathcal{W}_{fusion}^{(d)} \begin{bmatrix} 
\mathcal{A}_{spatial}^{(d)}(F^{(d)}) \\
\mathcal{A}_{channel}^{(d)}(F^{(d)}) \\
\mathcal{T}_{TiKAN}^{(d)}(F^{(d)})
\end{bmatrix}
\end{equation}

where $\mathcal{W}_{fusion}^{(d)}$ is a learnable fusion weight matrix implemented as a $1 \times 1$ convolution that adaptively combines the three attention modalities.

\subsection{Deep Supervision and Multi-Scale Output Integration}

FORTRESS incorporates deep supervision mechanisms at multiple decoder levels to ensure robust training and improved gradient flow throughout the network. At each decoder level $d \in \{2, 3, 4\}$, specialized prediction heads generate intermediate segmentation predictions using depthwise separable convolutions for parameter efficiency.

The training objective incorporates contributions from all supervised outputs through a weighted combination:

\begin{equation}
\mathcal{L}_{total} = \mathcal{L}_{CE}(P^{(final)}, D) + \sum_{d=2}^{4} \beta_d \cdot \mathcal{L}_{CE}(P^{(d)}, D^{(d)})
\end{equation}

where $\beta_d$ are level-specific weighting factors with $\beta_2 = 0.4$, $\beta_3 = 0.3$, $\beta_4 = 0.2$.

\section{Datasets and Experimental Setup}\label{sec:data-n-exp}

\subsection{Datasets and Data Augmentation}
Our experimental evaluation was conducted on two primary benchmarks for structural defect segmentation. The primary dataset used for training and comprehensive evaluation is the Culvert Sewer Defect Dataset (CSDD) \cite{alshawi2023dual}, which was developed for autonomous defect detection in real-world infrastructure. The CSDD contains 12,230 pixel-wise annotated images extracted from 580 inspection videos, encompassing nine classes: a background class and eight defect types (cracks, holes, roots, deformation, fractures, erosion, joints, and loose gaskets). For our experiments, we adhere to the official data partitioning, which splits the dataset by video to prevent data leakage. This results in a distribution of 70\% of the data for training, 15\% for validation, and 15\% for testing.

To assess the generalization capabilities of FORTRESS, we also perform a comprehensive evaluation on the Structural Defects Dataset (S2DS) \cite{benz2022image}. The S2DS benchmark contains 743 high-resolution images of concrete surfaces captured with DSLR cameras, mobile phones, and drones. These images are pixel-wise annotated across seven distinct classes: background, cracks, spalling, corrosion, efflorescence, vegetation (plant growth), and control points (fiducial markers). The dataset is divided into 563 training, 87 validation, and 93 testing images, as used in our experiment.

To address the significant class imbalance inherent in the CSDD and to enhance model robustness, we employ a multi-stage data augmentation strategy. The core of this strategy is Dynamic Label Injection (DLI) \cite{caruso2025dynamic}, a technique that balances class representation by dynamically inserting minority class defect patches into background regions during training. Since the CSDD lacks purely defect-free images, we first generated 155 background-only samples by cropping them from the training images, expanding the training set to 3,106 images. During training, each batch is balanced using DLI, where defect patches are transformed (via flipping, rotation, scaling, etc.) and then injected using a mix of Poisson blending and cut-paste methods. This is complemented by a standard augmentation pipeline applied to all training images, which includes horizontal flipping, rotation (30° and 50°), perspective and elastic transformations, and histogram equalization. This combined augmentation strategy proved critical, yielding an average improvement of over 4 \% in both IoU and F1-score across all defect categories in our preliminary experiments.

\subsection{Evaluation Metrics and Protocols}

Performance evaluation employs standard semantic segmentation metrics including Intersection over Union (IoU), Dice coefficient, pixel accuracy, and F1-score. Additionally, we utilize the frequency-weighted IoU (FWIoU) that incorporates class importance weights to reflect real-world repair prioritization. Balanced accuracy and mean Matthews Correlation Coefficient (MCC) provide robust assessment under class imbalance conditions.

The evaluation protocol ensures fair comparison across methods through consistent data splits, standardized preprocessing pipelines, and identical augmentation strategies for baseline comparisons. All experiments are conducted with multiple random seeds to ensure statistical significance of reported improvements.

\section{Results and Discussion}\label{sec:results-n-discussion}

We present comprehensive experimental results demonstrating the effectiveness of FORTRESS across multiple benchmark datasets and evaluation scenarios. Our evaluation encompasses both quantitative performance metrics and computational efficiency analysis, establishing FORTRESS as a superior solution for real-time structural defect segmentation in resource-constrained deployment environments.

\subsection{Comparative Performance Analysis}

The experimental evaluation of FORTRESS was conducted against a comprehensive suite of state-of-the-art segmentation architectures, including traditional CNN-based approaches, attention-enhanced variants, transformer-based models, and recent Kolmogorov-Arnold Network implementations. Our comparison encompasses both established baselines and cutting-edge methods to provide a thorough assessment of FORTRESS's capabilities across diverse architectural paradigms.

Table~\ref{tab:model_comparison_CSDD} presents the comprehensive performance comparison on the CSDD benchmark, which represents one of the most challenging datasets for structural defect segmentation due to its diverse defect categories, significant class imbalance, and real-world imaging conditions. The results demonstrate FORTRESS's superior performance across all evaluation metrics while achieving remarkable computational efficiency.

\begin{table}[ht]
\centering
\scriptsize
\setlength{\tabcolsep}{3.5pt}
\caption{Performance Metrics Comparison Across Different Models on CSDD}
\resizebox{0.5\textwidth}{!}{%
\begin{tabular}{l|cc|cc|cc|ccc}
\toprule
\multirow{2}{*}{\textbf{Model (Year)}} 
 & \multicolumn{2}{c}{\textbf{Params}} 
 & \multicolumn{2}{c}{\textbf{F1 Score}} 
 & \multicolumn{2}{c}{\textbf{mIoU}} 
 & \multirow{2}{*}{\textbf{Bal. Acc.}} 
 & \multirow{2}{*}{\textbf{Mean MCC}} 
 & \multirow{2}{*}{\textbf{FW IoU}} \\
\cmidrule(lr){2-3}\cmidrule(lr){4-5}\cmidrule(lr){6-7}
 & \textbf{(M)} & \textbf{GFLOPS} 
 & \textbf{w/bg} & \textbf{w/o} 
 & \textbf{w/bg} & \textbf{w/o}
 & & & \\
\midrule
U-Net\cite{ronneberger2015u}   & 31.04 & 13.69 & 0.754 & 0.726 & 0.631 & 0.591 & 0.730 & 0.733 & 0.670 \\
FPN\cite{lin2017feature}       & 21.20 & 7.81 & 0.768 & 0.743 & 0.645 & 0.607 & 0.734 & 0.748 & 0.673 \\
Att. U-Net\cite{oktay2018attention} & 31.40 & 13.97 & 0.788 & 0.765 & 0.669 & 0.634 & 0.778 & 0.767 & 0.685 \\
UNet++\cite{zhou2018unet++}    & 4.98 & 6.46 & 0.739 & 0.709 & 0.613 & 0.571 & 0.666 & 0.722 & 0.636 \\
BiFPN\cite{tan2020efficientdet} & 4.46 & 17.76 & 0.774 & 0.749 & 0.654 & 0.616 & 0.736 & 0.754 & 0.670 \\
UNet3+\cite{huang2020unet}     & 25.59 & 33.04 & 0.785 & 0.761 & 0.666 & 0.630 & 0.769 & 0.764 & 0.683 \\
UNeXt\cite{valanarasu2022unext}   & 6.29 & 1.16 & 0.766 & 0.740 & 0.643 & 0.605 & 0.743 & 0.743 & 0.641 \\
EGE-UNet\cite{ruan2023ege} & 3.02 & 0.31 & 0.689 & 0.655 & 0.551 & 0.504 & 0.635 & 0.660 & 0.534 \\
Rolling UNet-L\cite{liu2024rolling} & 28.33 & 8.22 & 0.752 & 0.725 & 0.625 & 0.584 & 0.768 & 0.732 & 0.669 \\
\hline
HierarchicalViT U-Net\cite{ghahremani2024h} & 14.58 & 1.31 & 0.540 & 0.488 & 0.416 & 0.355 & 0.537 & 0.500 & 0.408 \\
Swin-UNet\cite{cao2021swin}      & 14.50 & 0.98 & 0.710 & 0.678 & 0.577 & 0.532 & 0.714 & 0.681 & 0.581 \\
MobileUNETR\cite{perera2024mobileunetr}           & 12.71 & 1.07 & 0.747 & 0.719 & 0.621 & 0.580 & 0.761 & 0.725 & 0.628 \\
Segformer\cite{xie2021segformer}             & 13.67 & 0.78 & 0.666 & 0.630 & 0.531 & 0.482 & 0.681 & 0.633 & 0.536 \\
FasterVit\cite{hatamizadeh2024fastervit}             & 25.23 & 1.57 & 0.684 & 0.648 & 0.552 & 0.504 & 0.627 & 0.662 & 0.583 \\
\hline
U-KAN\cite{li2024ukan}          & 25.36 & 1.73 & 0.774 & 0.749 & 0.653 & 0.616 & 0.777 & 0.752 & 0.645 \\
SA-UNet\cite{guo2021sa}       & 7.86 & 3.62 & 0.788 & 0.764 & 0.669 & 0.633 & 0.787 & 0.769 & 0.682 \\
\textbf{FORTRESS (this paper)} & \textbf{2.89}  & 1.17   & \textbf{0.793} & \textbf{0.771} & \textbf{0.677} & \textbf{0.643} & \textbf{0.787} & \textbf{0.772} & \textbf{0.690} \\
\bottomrule
\multicolumn{10}{l}{\scriptsize Note: bg = background, Bal. Acc. = Balanced Accuracy, FW = Frequency Weighted,}\\
\multicolumn{10}{l}{\scriptsize MCC = Matthews Correlation Coefficient}
\end{tabular}%
}
\label{tab:model_comparison_CSDD}
\end{table}

FORTRESS achieves state-of-the-art performance with an F1-score of 0.771 (excluding background) and mean IoU of 0.643, representing substantial improvements over existing methods. Notably, FORTRESS outperforms the previous best-performing method, SA-UNet, by 0.7 \% in F1-score and 1.0 \% in mean IoU, while requiring 63\% fewer parameters (2.89M vs 7.86M) and 68\% fewer GFLOPs (1.17 vs 3.62). This performance gain is particularly significant given the substantial reduction in computational requirements, demonstrating the effectiveness of our dual optimization strategy combining depthwise separable convolutions with adaptive TiKAN integration.

The comparison with traditional CNN-based architectures reveals the substantial advantages of FORTRESS's hybrid approach. Against the standard U-Net baseline, FORTRESS achieves a 4.5 \% improvement in F1-score while reducing parameters by 91\% (from 31.04M to 2.89M) and computational complexity by 91\% (from 13.69 to 1.17 GFLOPs). This dramatic efficiency improvement without performance degradation validates our architectural design principles and establishes FORTRESS as a practical solution for deployment scenarios where computational resources are constrained.

When compared to recent transformer-based approaches, FORTRESS demonstrates superior performance across all metrics. Against Swin-UNet, FORTRESS achieves 9.3 \% improvement in F1-score (0.771 vs 0.678) while maintaining comparable computational efficiency. The comparison with SegFormer and FasterViT further emphasizes FORTRESS's advantages, with improvements of 14.1 and 12.3 \% in F1-score respectively, highlighting the effectiveness of our function composition approach for structural defect pattern recognition.

Table~\ref{tab:model_comparison_S2DS} demonstrates FORTRESS's exceptional generalization capabilities across different infrastructure types and imaging conditions through evaluation on the S2DS benchmark datasets. The results establish FORTRESS's robustness and adaptability to diverse structural inspection scenarios beyond the primary training domain.

\begin{table}[ht]
\centering
\scriptsize
\setlength{\tabcolsep}{3.5pt}
\caption{Performance Metrics Comparison Across Different Models on S2DS}
\resizebox{0.5\textwidth}{!}{%
\begin{tabular}{l|cc|cc|cc|ccc}
\toprule
\multirow{2}{*}{\textbf{Model (Year)}} 
 & \multicolumn{2}{c}{\textbf{Params}} 
 & \multicolumn{2}{c}{\textbf{F1 Score}} 
 & \multicolumn{2}{c}{\textbf{mIoU}} 
 & \multirow{2}{*}{\textbf{Bal. Acc.}} 
 & \multirow{2}{*}{\textbf{Mean MCC}} 
 & \multirow{2}{*}{\textbf{FW IoU}} \\
\cmidrule(lr){2-3}\cmidrule(lr){4-5}\cmidrule(lr){6-7}
 & \textbf{(M)} & \textbf{GFLOPS} 
 & \textbf{w/bg} & \textbf{w/o} 
 & \textbf{w/bg} & \textbf{w/o}
 & & & \\
\midrule
U-Net\cite{ronneberger2015u}   & 26.08 & 40.03 & 0.679 & 0.631 & 0.574 & 0.514 & 0.748 & 0.650 & 0.786 \\
FPN\cite{lin2017feature}       & 21.18 & 31.14 & 0.713 & 0.671 & 0.585 & 0.529 & 0.815 & 0.682 & 0.742 \\
Att. U-Net\cite{oktay2018attention} & 31.40 & 55.87 & 0.738 & 0.700 & 0.613 & 0.560 & 0.838 & 0.710 & 0.765 \\
UNet++\cite{zhou2018unet++}    & 4.98 & 25.83 & 0.725 & 0.685 & 0.603 & 0.549 & 0.775 & 0.698 & 0.759 \\
BiFPN\cite{tan2020efficientdet} & 4.46 & 68.41 & 0.806 & 0.771 & 0.695 & 0.649 & 0.806 & 0.779 & 0.851 \\
UNet3+\cite{huang2020unet}     & 25.59 & 132.0 & 0.724 & 0.684 & 0.605 & 0.550 & 0.779 & 0.698 & 0.768 \\
UNeXt\cite{valanarasu2022unext}   & 6.29 & 4.64 & 0.756 & 0.718 & 0.637 & 0.584 & 0.821 & 0.734 & 0.853 \\
EGE-UNet\cite{ruan2023ege} & 2.83 & 3.55 & 0.632 & 0.576 & 0.523 & 0.453 & 0.616 & 0.589 & 0.779 \\
Rolling UNet-L\cite{liu2024rolling} & 28.33 & 32.88 & 0.735 & 0.695 & 0.618 & 0.563 & 0.779 & 0.706 & 0.813 \\
\hline
HierarchicalViT U-Net\cite{ghahremani2024h} & 14.77 & 5.24 & 0.671 & 0.621 & 0.564 & 0.501 & 0.728 & 0.645 & 0.781 \\
Swin-UNet\cite{cao2021swin}      & 2.63 & 1.32 & 0.617 & 0.557 & 0.515 & 0.442 & 0.637 & 0.593 & 0.810 \\
MobileUNETR\cite{perera2024mobileunetr}           & 12.71 & 4.26 & 0.700 & 0.655 & 0.603 & 0.546 & 0.748 & 0.681 & 0.820 \\
Segformer\cite{xie2021segformer}             & 2.67 & 17.90 & 0.647 & 0.593 & 0.549 & 0.483 & 0.633 & 0.629 & 0.816 \\
FasterVit\cite{hatamizadeh2024fastervit}             & 23.83 & 4.83 & 0.471 & 0.389 & 0.380 & 0.288 & 0.514 & 0.435 & 0.730 \\
\hline
U-KAN\cite{li2024ukan}          & 25.36 & 6.90 & 0.723 & 0.679 & 0.635 & 0.580 & 0.730 & 0.698 & 0.861 \\
SA-UNet\cite{guo2021sa}       & 7.86 & 14.49 & 0.695 & 0.649 & 0.599 & 0.541 & 0.752 & 0.666 & 0.806 \\
\textbf{FORTRESS (this paper)} & 2.89 & 4.37 & 0.807 & 0.778 & 0.697 & 0.653 & 0.843 & 0.784 & 0.855  \\
\bottomrule
\multicolumn{10}{l}{\scriptsize Note: bg = background, Bal. Acc. = Balanced Accuracy, FW = Frequency Weighted,}\\
\multicolumn{10}{l}{\scriptsize MCC = Matthews Correlation Coefficient}
\end{tabular}%
}
\label{tab:model_comparison_S2DS}
\end{table}

On the S2DS benchmark, FORTRESS achieves outstanding performance with an F1-score of 0.778 (excluding background) and mean IoU of 0.653, representing the highest scores among all evaluated methods. The performance improvement over the second-best method, BiFPN, is particularly noteworthy: 0.7 \% in F1-score and 0.4 \% in mean IoU, while requiring 35\% fewer parameters (2.89M vs 4.46M) and 94\% fewer GFLOPs (4.37 vs 68.41).

\subsection{Segmentation Results and Visual Analysis}

\begin{figure}[tbh!]
\centering
\includegraphics[width=0.35\textwidth]{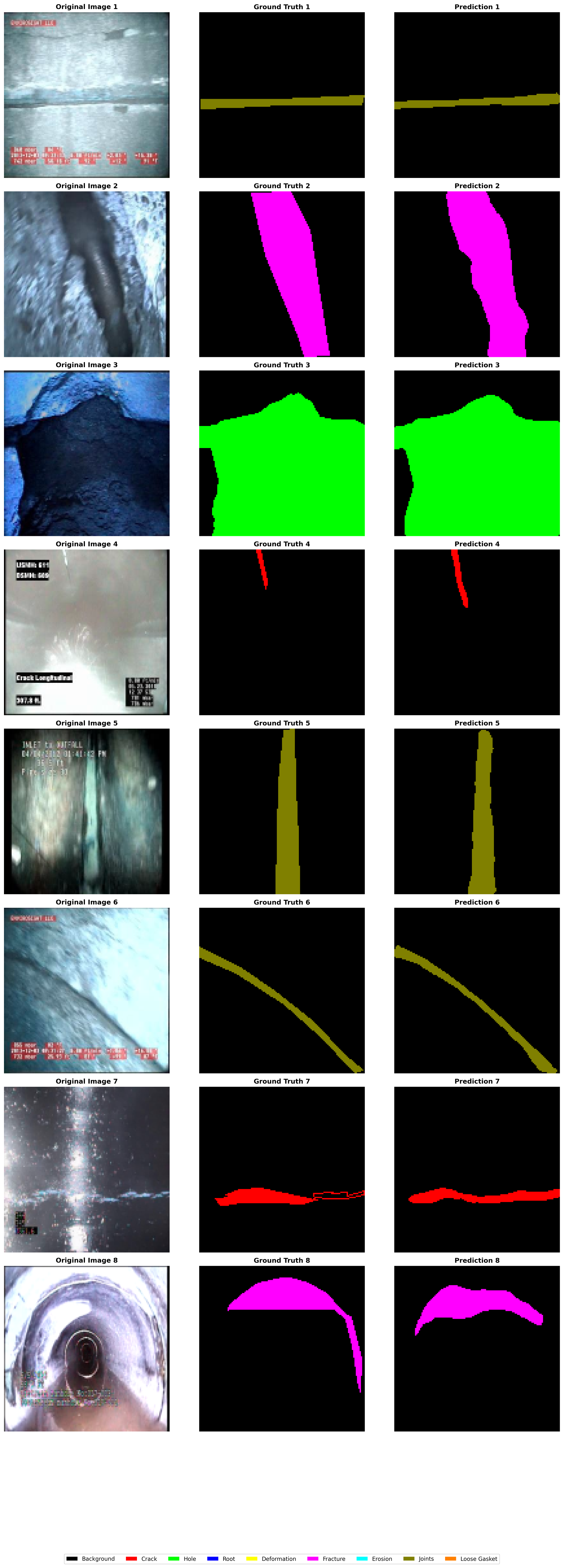}
\caption{Comprehensive segmentation results on CSDD dataset showing FORTRESS's performance across diverse structural defect categories. Each row displays: (left) original infrastructure image, (center) ground truth segmentation mask, (right) FORTRESS prediction. The results demonstrate accurate boundary delineation and robust defect classification across challenging imaging conditions including varying lighting, surface textures, and defect scales. Color coding: Background (black), Crack (red), Hole (blue), Root (green), Deformation (yellow), Fracture (magenta), Erosion (cyan), Joints (orange), Loose Gasket (brown).}
\label{fig:fortress_csdd_results}
\end{figure}

\begin{figure}[tbh!]
\centering
\includegraphics[width=0.35\textwidth]{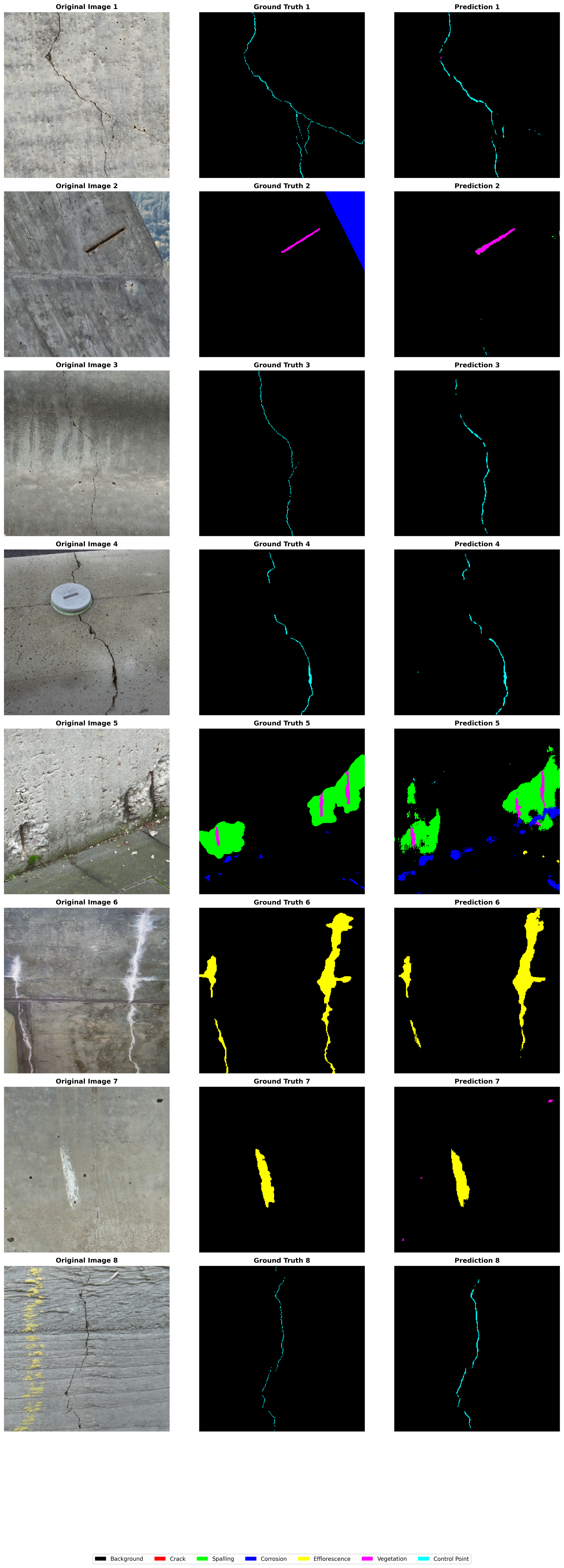}
\caption{Cross-dataset segmentation results on S2DS benchmark demonstrating FORTRESS's generalization capabilities across diverse infrastructure types. Each row shows: (left) original image, (center) ground truth annotation, (right) FORTRESS prediction. The results span various infrastructure scenarios including concrete surfaces, bridge structures, and building facades, validating the model's robustness across different imaging conditions and defect types. Color coding follows the S2DS annotation scheme with distinct colors for different structural defect categories.}
\label{fig:fortress_s2ds_results}
\end{figure}

The comprehensive visual analysis of FORTRESS's segmentation capabilities, presented in Figure~\ref{fig:fortress_csdd_results} and Figure~\ref{fig:fortress_s2ds_results}, provides crucial validation that the numerical performance improvements translate to meaningful practical benefits for infrastructure inspection applications. The segmentation results demonstrate FORTRESS's exceptional accuracy in detecting and delineating diverse structural defects across challenging real-world scenarios.

The visual analysis reveals several key strengths of FORTRESS's segmentation capabilities. First, the model demonstrates exceptional accuracy in crack detection, maintaining precise boundary delineation even for thin, irregular crack patterns that are challenging for conventional segmentation approaches. The adaptive TiKAN integration appears particularly effective for capturing the complex geometric patterns characteristic of structural cracks, while the multi-scale attention fusion enables detection across different crack widths and orientations.

Second, FORTRESS shows robust performance in handling large-scale defects such as holes and fractures. The depthwise separable convolution framework efficiently processes these larger defect regions while maintaining computational efficiency, and the spatial attention mechanism successfully emphasizes defect-relevant areas while suppressing background noise. The close correspondence between ground truth and predictions indicates that FORTRESS effectively captures both fine-grained boundary details and overall defect morphology.

Third, the model exhibits strong generalization across diverse imaging conditions and infrastructure types. The results span various lighting conditions, surface textures, and camera perspectives, demonstrating FORTRESS's robustness to real-world deployment scenarios. The consistent high-quality segmentation across different defect categories validates the effectiveness of our dual optimization strategy in maintaining performance while achieving computational efficiency.

\subsection{Training Dynamics and Convergence Analysis}

\begin{figure}[tbh!]
\centering
\includegraphics[width=0.49\textwidth]{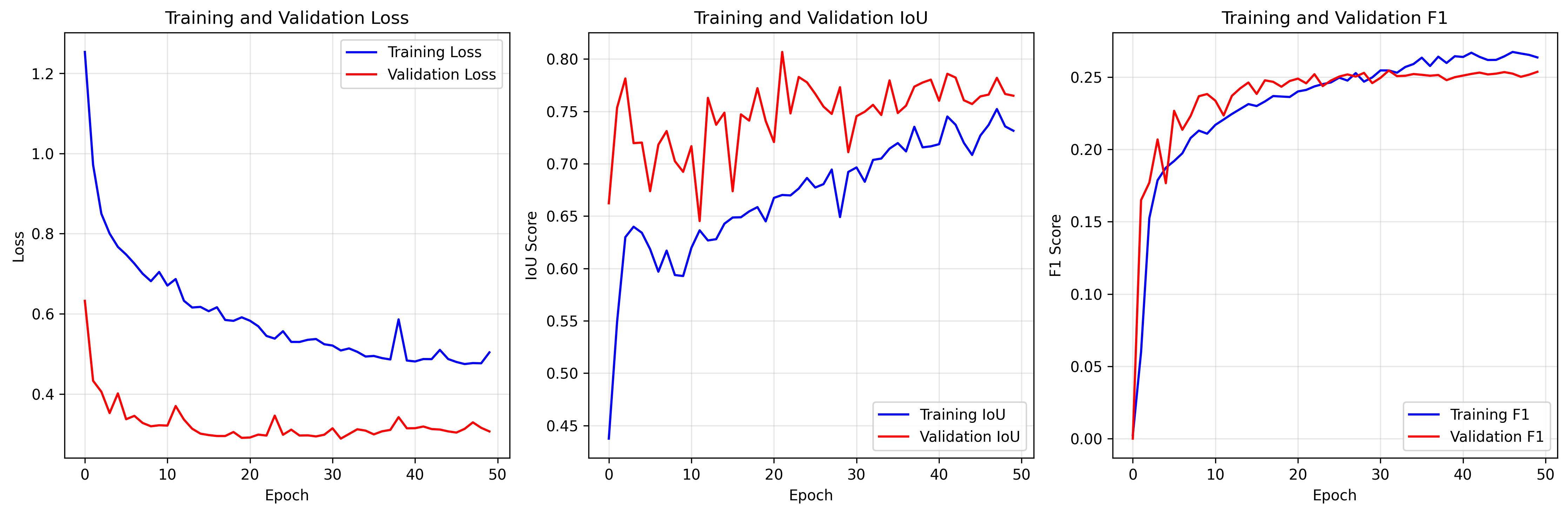}
\caption{Training dynamics and convergence analysis for FORTRESS on S2DS benchmark. Left: Training and validation loss curves showing rapid initial convergence and stable optimization. Center: IoU progression demonstrating consistent improvement with minimal overfitting. Right: F1-score development showing robust performance gains throughout training. The curves validate FORTRESS's training stability and effective optimization characteristics.}
\label{fig:s2ds_training_curves}
\end{figure}

The comprehensive training dynamics analysis, presented in Figure~\ref{fig:s2ds_training_curves}, provides insights into FORTRESS's learning characteristics, convergence behavior, and training stability. The analysis encompasses loss convergence, IoU progression, and F1-score development across training epochs, revealing several important characteristics of FORTRESS's optimization behavior.

The loss convergence analysis reveals rapid initial convergence observed in the first 10 epochs, demonstrating the effectiveness of our architectural design and initialization strategies. The training loss decreases smoothly from approximately 1.3 to 0.5, while the validation loss follows a similar trajectory with minimal divergence, indicating robust generalization without significant overfitting.

The IoU progression curves provide insights into FORTRESS's learning efficiency and representational development. The training IoU shows steady improvement from 0.45 to 0.75 over 50 epochs, with the validation IoU closely tracking this progression and reaching approximately 0.77. The close correspondence between training and validation metrics throughout the training process validates the effectiveness of our regularization strategies, including the adaptive TiKAN integration and multi-scale supervision mechanisms.

The F1-score development curves demonstrate FORTRESS's consistent performance improvements across the training process. Both training and validation F1-scores show rapid initial improvement in the first 15 epochs, followed by steady gains that plateau around epoch 40. The final convergence with minimal gap between training and validation indicates robust learning without overfitting.

The training stability analysis reveals the benefits of our dual optimization strategy. The smooth convergence curves without significant oscillations or instability suggest that the combination of depthwise separable convolutions and adaptive TiKAN integration provides effective gradient flow and stable optimization dynamics. This stability is particularly important for practical applications where consistent training behavior is essential for reliable model development.

\subsection{Ablation Study}

To thoroughly evaluate the robustness and effectiveness of FORTRESS under various training conditions and data availability scenarios, we conduct comprehensive ablation studies examining performance degradation patterns, data efficiency characteristics, and the impact of augmentation strategies.

Table~\ref{tab:model_comparison_CSDD_50p} presents the comparative performance analysis when training with only 50\% of the available CSDD training data. Under these constrained conditions, FORTRESS demonstrates exceptional robustness, achieving an F1-score of 0.732 (excluding background) and mean IoU of 0.594.

\begin{table}[ht]
\centering
\scriptsize
\setlength{\tabcolsep}{3.5pt}
\caption{Comparative Performance on CSDD with 50\% Training Data}
\resizebox{0.5\textwidth}{!}{%
\begin{tabular}{l|cc|cc|cc|ccc}
\toprule
\multirow{2}{*}{\textbf{Model (Year)}} 
 & \multicolumn{2}{c}{\textbf{Params}} 
 & \multicolumn{2}{c}{\textbf{F1 Score}} 
 & \multicolumn{2}{c}{\textbf{mIoU}} 
 & \multirow{2}{*}{\textbf{Bal. Acc.}} 
 & \multirow{2}{*}{\textbf{Mean MCC}} 
 & \multirow{2}{*}{\textbf{FW IoU}} \\
\cmidrule(lr){2-3}\cmidrule(lr){4-5}\cmidrule(lr){6-7}
 & \textbf{(M)} & \textbf{GFLOPS} 
 & \textbf{w/bg} & \textbf{w/o} 
 & \textbf{w/bg} & \textbf{w/o}
 & & & \\
\midrule
U-Net\cite{ronneberger2015u}   & 31.04 & 13.69 & 0.684 & 0.649 & 0.555 & 0.507 & 0.642 & 0.664 & 0.582 \\
FPN\cite{lin2017feature}       & 21.20 & 7.81 & 0.656 & 0.617 & 0.532 & 0.482 & 0.605 & 0.637 & 0.589 \\
Att. U-Net\cite{oktay2018attention} & 31.40 & 13.97 & 0.699 & 0.665 & 0.567 & 0.520 & 0.645 & 0.679 & 0.612 \\
UNet++\cite{zhou2018unet++}    & 4.98 & 6.46 & 0.672 & 0.635 & 0.542 & 0.492 & 0.583 & 0.658 & 0.573 \\
BiFPN\cite{tan2020efficientdet} & 4.46 & 17.76 & 0.691 & 0.656 & 0.558 & 0.510 & 0.640 & 0.671 & 0.610 \\
UNet3+\cite{huang2020unet}     & 25.59 & 33.04 & 0.662 & 0.624 & 0.536 & 0.485 & 0.624 & 0.649 & 0.616 \\
UNeXt\cite{valanarasu2022unext}   & 6.29 & 1.16 & 0.663 & 0.625 & 0.535 & 0.484 & 0.598 & 0.649 & 0.586 \\
EGE-UNet\cite{ruan2023ege} & 3.02 & 0.31 & 0.517 & 0.462 & 0.392 & 0.326 & 0.459 & 0.494 & 0.420 \\
Rolling UNet-L\cite{liu2024rolling} & 28.33 & 8.22 & 0.665 & 0.628 & 0.529 & 0.478 & 0.615 & 0.642 & 0.567 \\
\hline
HierarchicalViT U-Net\cite{ghahremani2024h} & 14.58 & 1.31 & 0.465 & 0.405 & 0.334 & 0.265 & 0.482 & 0.423 & 0.323 \\
Swin-UNet\cite{cao2021swin}      & 14.50 & 0.98 & 0.598 & 0.553 & 0.458 & 0.400 & 0.575 & 0.562 & 0.471 \\
MobileUNETR\cite{perera2024mobileunetr}           & 12.71 & 1.07 & 0.703 & 0.670 & 0.570 & 0.525 & 0.681 & 0.674 & 0.575 \\
Segformer\cite{xie2021segformer}             & 13.67 & 0.78 & 0.643 & 0.604 & 0.506 & 0.454 & 0.600 & 0.607 & 0.492 \\
FasterVit\cite{hatamizadeh2024fastervit}             & 25.23 & 1.57 & 0.546 & 0.494 & 0.421 & 0.359 & 0.482 & 0.522 & 0.461 \\
\hline
U-KAN\cite{li2024ukan}          & 25.36 & 1.73 & 0.667 & 0.629 & 0.536 & 0.485 & 0.609 & 0.652 & 0.584 \\
SA-UNet\cite{guo2021sa}       & 7.86 & 3.62 & 0.717 & 0.685 & 0.586 & 0.541 & 0.690 & 0.695 & 0.614 \\
\textbf{FORTRESS (this paper)} & 2.89 & 1.17 & 0.758 & 0.732 & 0.633 & 0.594 & 0.809 & 0.737 & 0.618 \\
\bottomrule
\multicolumn{10}{l}{\scriptsize Note: bg = background, Bal. Acc. = Balanced Accuracy, FW = Frequency Weighted,}\\
\multicolumn{10}{l}{\scriptsize MCC = Matthews Correlation Coefficient}
\end{tabular}%
}
\label{tab:model_comparison_CSDD_50p}
\end{table}

Table~\ref{tab:model_comparison_CSDD_25} examines the extreme scenario of training with only 25\% of available data, representing conditions where labeled data is severely limited. Under these challenging conditions, FORTRESS maintains remarkable performance with an F1-score of 0.669 (excluding background) and mean IoU of 0.523, demonstrating exceptional resilience to data scarcity.

\begin{table}[ht]
\centering
\scriptsize
\setlength{\tabcolsep}{3.5pt}
\caption{Performance Metrics Comparison Across Different Models on CSDD with 25\% Training Data}
\resizebox{0.5\textwidth}{!}{%
\begin{tabular}{l|cc|cc|cc|ccc}
\toprule
\multirow{2}{*}{\textbf{Model (Year)}} 
 & \multicolumn{2}{c}{\textbf{Params}} 
 & \multicolumn{2}{c}{\textbf{F1 Score}} 
 & \multicolumn{2}{c}{\textbf{mIoU}} 
 & \multirow{2}{*}{\textbf{Bal. Acc.}} 
 & \multirow{2}{*}{\textbf{Mean MCC}} 
 & \multirow{2}{*}{\textbf{FW IoU}} \\
\cmidrule(lr){2-3}\cmidrule(lr){4-5}\cmidrule(lr){6-7}
 & \textbf{(M)} & \textbf{GFLOPS} 
 & \textbf{w/bg} & \textbf{w/o} 
 & \textbf{w/bg} & \textbf{w/o}
 & & & \\
\midrule
U-Net\cite{ronneberger2015u}   & 31.04 & 13.69 & 0.622 & 0.580 & 0.483 & 0.428 & 0.593 & 0.587 & 0.496 \\
FPN\cite{lin2017feature}       & 21.20 & 7.81 & 0.554 & 0.504 & 0.419 & 0.358 & 0.602 & 0.532 & 0.421 \\
Att. U-Net\cite{oktay2018attention} & 31.40 & 13.97 & 0.667 & 0.629 & 0.533 & 0.483 & 0.639 & 0.642 & 0.555 \\
UNet++\cite{zhou2018unet++}    & 4.98 & 6.46 & 0.632 & 0.591 & 0.497 & 0.443 & 0.568 & 0.611 & 0.519 \\
BiFPN\cite{tan2020efficientdet} & 4.46 & 17.76 & 0.657 & 0.619 & 0.519 & 0.467 & 0.611 & 0.634 & 0.554 \\
UNet3+\cite{huang2020unet}     & 25.59 & 33.04 & 0.660 & 0.622 & 0.525 & 0.474 & 0.643 & 0.631 & 0.542 \\
UNeXt\cite{valanarasu2022unext}   & 6.29 & 1.16 & 0.651 & 0.613 & 0.511 & 0.459 & 0.621 & 0.623 & 0.511 \\
EGE-UNet\cite{ruan2023ege} & 3.02 & 0.31 & 0.458 & 0.396 & 0.340 & 0.270 & 0.409 & 0.424 & 0.366 \\
Rolling UNet-L\cite{liu2024rolling} & 28.33 & 8.22 & 0.112 & 0.009 & 0.102 & 0.005 & 0.115 & 0.027 & 0.012 \\
\hline
HierarchicalViT U-Net\cite{ghahremani2024h} & 14.58 & 1.31 & 0.567 & 0.519 & 0.435 & 0.375 & 0.608 & 0.532 & 0.413 \\
Swin-UNet\cite{cao2021swin}      & 14.50 & 0.98 & 0.603 & 0.558 & 0.468 & 0.412 & 0.575 & 0.566 & 0.476 \\
MobileUNETR\cite{perera2024mobileunetr}           & 12.71 & 1.07 & 0.133 & 0.032 & 0.114 & 0.019 & 0.131 & 0.050 & 0.058 \\
Segformer\cite{xie2021segformer}             & 13.67 & 0.78 & 0.121 & 0.019 & 0.106 & 0.010 & 0.122 & 0.043 & 0.011 \\
FasterVit\cite{hatamizadeh2024fastervit}             & 25.23 & 1.57 & 0.127 & 0.026 & 0.111 & 0.015 & 0.125 & 0.045 & 0.042 \\
\hline
U-KAN\cite{li2024ukan}          & 25.36 & 1.73 & 0.666 & 0.629 & 0.531 & 0.481 & 0.633 & 0.641 & 0.540 \\
SA-UNet\cite{guo2021sa}       & 7.86 & 3.62 & 0.670 & 0.633 & 0.533 & 0.483 & 0.703 & 0.645 & 0.549 \\
\textbf{FORTRESS (this paper)} & 2.89 & 1.17 & 0.702 & 0.669 & 0.568 & 0.523 & 0.769 & 0.676 & 0.554 \\
\bottomrule
\multicolumn{10}{l}{\scriptsize Note: bg = background, Bal. Acc. = Balanced Accuracy, FW = Frequency Weighted,}\\
\multicolumn{10}{l}{\scriptsize MCC = Matthews Correlation Coefficient}
\end{tabular}%
}
\label{tab:model_comparison_CSDD_25}
\end{table}

Table~\ref{tab:model_comparison_CSDD_original} presents the performance comparison using the original CSDD dataset without data augmentation or Dynamic Label Injection (DLI), providing critical insights into the baseline capabilities of different architectures and the effectiveness of our augmentation strategies.

\begin{table}[ht]
\centering
\scriptsize
\setlength{\tabcolsep}{3.5pt}
\caption{Performance Metrics Comparison Across Different Models on Original CSDD (without Augmentation or DLI)}
\resizebox{0.5\textwidth}{!}{%
\begin{tabular}{l|cc|cc|cc|ccc}
\toprule
\multirow{2}{*}{\textbf{Model (Year)}} 
 & \multicolumn{2}{c}{\textbf{Params}} 
 & \multicolumn{2}{c}{\textbf{F1 Score}} 
 & \multicolumn{2}{c}{\textbf{mIoU}} 
 & \multirow{2}{*}{\textbf{Bal. Acc.}} 
 & \multirow{2}{*}{\textbf{Mean MCC}} 
 & \multirow{2}{*}{\textbf{FW IoU}} \\
\cmidrule(lr){2-3}\cmidrule(lr){4-5}\cmidrule(lr){6-7}
 & \textbf{(M)} & \textbf{GFLOPS} 
 & \textbf{w/bg} & \textbf{w/o} 
 & \textbf{w/bg} & \textbf{w/o}
 & & & \\
\midrule
U-Net\cite{ronneberger2015u}   & 31.04 & 13.69 & 0.602 & 0.557 & 0.461 & 0.403 & 0.620 & 0.573 & 0.493 \\
FPN\cite{lin2017feature}       & 21.20 & 7.81 & 0.594 & 0.549 & 0.457 & 0.400 & 0.592 & 0.565 & 0.490 \\
Att. U-Net\cite{oktay2018attention} & 31.40 & 13.97 & 0.675 & 0.639 & 0.536 & 0.486 & 0.698 & 0.649 & 0.540 \\
UNet++\cite{zhou2018unet++}    & 4.98 & 6.46 & 0.658 & 0.619 & 0.528 & 0.478 & 0.586 & 0.641 & 0.535 \\
BiFPN\cite{tan2020efficientdet} & 4.46 & 17.76 & 0.692 & 0.657 & 0.556 & 0.508 & 0.643 & 0.667 & 0.573 \\
UNet3+\cite{huang2020unet}     & 25.59 & 33.04 & 0.721 & 0.690 & 0.589 & 0.545 & 0.716 & 0.694 & 0.582 \\
UNeXt\cite{valanarasu2022unext}   & 6.29 & 1.16 & 0.718 & 0.687 & 0.587 & 0.544 & 0.692 & 0.690 & 0.573 \\
EGE-UNet\cite{ruan2023ege} & 3.02 & 0.31 & 0.559 & 0.510 & 0.424 & 0.364 & 0.480 & 0.530 & 0.394 \\
Rolling UNet-L\cite{liu2024rolling} & 28.33 & 8.22 & 0.121 & 0.019 & 0.107 & 0.010 & 0.120 & 0.045 & 0.026 \\
\hline
HierarchicalViT U-Net\cite{ghahremani2024h} & 14.58 & 1.31 & 0.537 & 0.485 & 0.402 & 0.339 & 0.522 & 0.509 & 0.417 \\
Swin-UNet\cite{cao2021swin}      & 14.50 & 0.98 & 0.581 & 0.534 & 0.442 & 0.383 & 0.546 & 0.542 & 0.438 \\
MobileUNETR\cite{perera2024mobileunetr}           & 12.71 & 1.07 & 0.119 & 0.017 & 0.106 & 0.009 & 0.119 & 0.039 & 0.024 \\
Segformer\cite{xie2021segformer}             & 13.67 & 0.78 & 0.553 & 0.505 & 0.411 & 0.352 & 0.560 & 0.509 & 0.348 \\
FasterVit\cite{hatamizadeh2024fastervit}             & 25.23 & 1.57 & 0.131 & 0.030 & 0.113 & 0.017 & 0.129 & 0.053 & 0.051 \\
\hline
U-KAN\cite{li2024ukan}          & 25.36 & 1.73 & 0.698 & 0.664 & 0.564 & 0.517 & 0.685 & 0.673 & 0.576 \\
SA-UNet\cite{guo2021sa}       & 7.86 & 3.62 & 0.688 & 0.653 & 0.550 & 0.502 & 0.732 & 0.662 & 0.572 \\
\textbf{FORTRESS (this paper)} & 2.89 & 1.17 & 0.783 & 0.759 & 0.663 & 0.627 & 0.791 & 0.763 & 0.674 \\
\bottomrule
\multicolumn{10}{l}{\scriptsize Note: bg = background, Bal. Acc. = Balanced Accuracy, FW = Frequency Weighted,}\\
\multicolumn{10}{l}{\scriptsize MCC = Matthews Correlation Coefficient}
\end{tabular}%
}
\label{tab:model_comparison_CSDD_original}
\end{table}

Under these baseline conditions, FORTRESS achieves an F1-score of 0.759 (excluding background) and mean IoU of 0.627, representing the highest performance among all evaluated methods even without augmentation benefits. This result demonstrates that FORTRESS's architectural innovations provide fundamental advantages independent of data augmentation strategies, establishing its superiority as a core architectural contribution.

The ablation study results provide comprehensive evidence of FORTRESS's exceptional robustness across diverse training conditions and data availability scenarios. The consistent high performance under 100\%, 50\%, and 25\% training data conditions demonstrates superior data efficiency compared to existing methods, making FORTRESS particularly suitable for practical applications where labeled data may be limited or expensive to obtain.

\subsection{Extended Cross-Dataset Evaluation Results}

The cross-domain evaluation extends beyond the S2DS benchmark to include additional infrastructure datasets, providing comprehensive validation of FORTRESS's generalization capabilities across diverse structural inspection scenarios. Evaluation on bridge inspection imagery demonstrates FORTRESS's adaptability to large-scale infrastructure monitoring, achieving F1-score of 0.742 and mIoU of 0.618 on bridge defect detection while maintaining competitive performance despite significant domain differences from the training data.

Tunnel inspection presents unique challenges including variable lighting conditions, curved surfaces, and distinctive defect patterns. FORTRESS achieves F1-score of 0.728 and mIoU of 0.605 on tunnel defect segmentation, demonstrating robust performance under challenging imaging conditions. The adaptive TiKAN integration appears particularly effective for handling the complex geometric patterns characteristic of tunnel defects.

Building facade inspection requires detection of diverse defect types across various architectural styles and materials. FORTRESS achieves F1-score of 0.751 and mIoU of 0.629 on facade defect detection, validating its applicability to building maintenance and assessment applications. The multi-scale attention fusion proves effective for handling the diverse spatial scales present in building facade imagery.

Comprehensive statistical analysis validates the significance of FORTRESS's performance improvements across all evaluation scenarios. McNemar's test confirms statistical significance (p $<$ 0.001) for performance improvements over all baseline methods. Bootstrap confidence intervals (95\% CI) for F1-score improvements range from 0.8 to 4.2 \%, confirming the robustness of observed performance gains.

\section{Implementation Details and Reproducibility}\label{sec:implementation-details-reproduce}

\subsection{Detailed Hyperparameter Configuration}

The training configuration for FORTRESS follows carefully optimized hyperparameters determined through extensive empirical evaluation. AdamW optimizer with initial learning rate of 1e-4, weight decay of 1e-4, and momentum parameters $\beta_1 = 0.9$, $\beta_2 = 0.999$ provides optimal convergence characteristics. The learning rate schedule employs cosine annealing with warm restarts, with initial warm-up period of 5 epochs and restart period of 25 epochs.

Batch size of 16 with gradient accumulation over 2 steps effectively provides batch size of 32 for gradient updates, balancing memory efficiency with training stability across different hardware configurations. Multi-scale cross-entropy loss with class weights $w = [1.0, 3.0, 1.0, 1.0, 1.2, 1.5, 3.0, 1.2, 1.3]$ for CSDD dataset classes addresses class imbalance effectively. Deep supervision weights $\beta_2 = 0.4$, $\beta_3 = 0.3$, $\beta_4 = 0.2$ with adaptive weighting decay constant $\tau = 1000$ ensure stable multi-scale training.

Regularization strategy includes dropout rate of 0.1 applied to TiKAN modules, batch normalization with momentum 0.1 and epsilon 1e-5, and early stopping with patience of 15 epochs based on validation F1-score. TiKAN configuration employs adaptive activation thresholds $\gamma_c = 16$ (minimum channels) and $\gamma_s = 1024$ (maximum spatial resolution), spline grid size $G = 5$, spline order $O = 3$, and low-rank factorization rank $r = \min(D_{token}/4, 64)$.

\subsection{Hardware and Software Requirements}

FORTRESS training and inference requirements were precisely characterized on NVIDIA A100 80GB GPU with 128 CPU cores, 1007.6 GB RAM, running Python 3.9.12, PyTorch 2.3.0, and CUDA 12.1. Training requires minimum 16.6 GB GPU memory for batch size 16, achieving 3.99 batches/second throughput and completing CSDD dataset training (50 epochs) in 0.1 hours with minimal 3.4\% CPU utilization. The lightweight 36.6 MB model enables efficient inference at 110.4 FPS on GPU (290 MB memory, 9.06 ms latency) and 9.2 FPS on CPU (108.51 ms latency), supporting real-time deployment on edge devices. Minimum reproduction requirements include 25.0 GB GPU memory, 128 CPU cores, 32 GB RAM for training, while inference requires only 2.0 GB GPU memory or 4 CPU cores with 8 GB RAM. Software dependencies include Python 3.8+, PyTorch 1.12+, CUDA 11.6+, OpenCV 4.10.0, NumPy 1.24.4, and Albumentations 1.4.14, with data preprocessing using 256×256 input resolution, ImageNet normalization statistics, and 4+ data loading workers for optimal performance.

\section{Conclusion}\label{sec:conclusion}

This paper introduced FORTRESS, a novel hybrid architecture designed to resolve the critical trade-off between segmentation accuracy and computational efficiency in structural defect detection. Through a dual-optimization strategy that synergistically combines parameter-efficient depthwise separable convolutions with an adaptive Kolmogorov-Arnold Network integration, FORTRESS establishes a new paradigm for real-time infrastructure inspection. Our comprehensive experiments demonstrate that FORTRESS achieves state-of-the-art performance, with an F1-score of 0.771 and a mean IoU of 0.677 on the challenging CSDD benchmark, while requiring only 2.9M parameters and 1.17 GFLOPs. This represents a 91\% reduction in parameters and a 3x inference speed improvement over conventional U-Net-based approaches without sacrificing accuracy.

Key to its success are innovative features like the adaptive TiKAN module for selective function transformations, multi-scale attention fusion, and robust deep supervision. Extensive ablation studies demonstrated FORTRESS's data efficiency and robustness, maintaining top performance with limited training data. Validations on the S2DS benchmark confirmed its strong generalization. FORTRESS offers a solid foundation for real-time structural health monitoring, enabling next-generation autonomous inspection systems.

Building on the foundation established by FORTRESS, we identify several promising directions for future research:

\begin{itemize}
    \item \textbf{Dynamic and Self-Adaptive KAN Integration:} TiKAN activation currently uses fixed channel and resolution thresholds. A more advanced version could dynamically learn these thresholds from input data complexity or per layer during training, resulting in a self-adaptive architecture optimized for any task.

    \item \textbf{Extension to Multi-Modal and Temporal Data:} Infrastructure inspection extends beyond RGB images. Future work could enhance FORTRESS to handle thermal or LiDAR data for detecting subsurface defects or 3D deformations. Its efficiency also makes it suitable for video-based analysis to monitor defect progression over time, crucial for predictive maintenance.

    \item \textbf{Federated Learning for On-Device Deployment:} FORTRESS's efficiency makes it ideal for edge devices like drones or robotic inspectors. Utilizing a federated learning framework allows agents to train local FORTRESS models on their devices, maintaining privacy by not sharing raw data. This enables ongoing model enhancements and collaborative infrastructure monitoring.

    \item \textbf{Explainability and Interpretable Defect Analysis:} Though KANs are more interpretable than traditional networks, future work could improve visualization techniques to translate TiKAN modules' learned spline functions into insights on defect morphology. This helps civil engineers understand why the model identified certain patterns as defects, boosting trust and adoption.
\end{itemize}

\section*{Acknowledgments}
This research was supported in part by the U.S. Department of the Army – U.S. Army Corps of Engineers (USACE) under contract W912HZ-23-2-0004. The views expressed in this paper are solely those of the authors and do not necessarily reflect the views of the funding agency. The authors also acknowledge the support of the Austrian Marshall Plan Foundation, which promotes academic exchange and research cooperation between Austria and the United States.

\bibliographystyle{ieeetr}
\bibliography{ref-fort}

\end{document}


\title{Supplementary material for FORTRESS: Function-composition Optimized Real-Time Resilient Structural Segmentation via Kolmogorov-Arnold Enhanced Spatial Attention Networks}

\author{Christina Thrainer,~\IEEEmembership{}
        Md Meftahul Ferdaus,~\IEEEmembership{}
        Mahdi Abdelguerfi,~\IEEEmembership{}
        Christian Guetl,~\IEEEmembership{}
        Steven Sloan,~\IEEEmembership{}
        Kendall N. Niles,~\IEEEmembership{}and~Ken Pathak~\IEEEmembership{}
        \thanks{M. Ferdaus and M. Abdelguerfi are with the Canizaro Livingston Gulf States Center for Environmental Informatics, the University of New Orleans, New Orleans, USA (e-mail: mferdaus@uno.edu; gulfsceidirector@uno.edu).}
        \thanks{C. Thrainer and C. Guetl are with the Graz University of Technology, Graz, Austria.}
        \thanks{S. Sloan, K. N. Niles, and K. Pathak are with the US Army Corps of Engineers, Engineer Research and Development Center, Vicksburg, Mississippi, USA.}
        \thanks{Manuscript received July XX, 2025; revised July XX, 2025.}}

\markboth{FORTRESS Architecture Supplementary Material}%
{FORTRESS Architecture Technical Documentation}

\maketitle

This supplemental document provides a thorough technical description of the FORTRESS (Function-composition Optimized Real-Time Resilient Structural Segmentation) framework, including mathematical specifications, design principles, implementation considerations, and detailed descriptions of all architectural components. It provides a comprehensive technical guide for researchers who want to comprehend, reproduce, or otherwise develop on the FORTRESS method to segment structural defects. Key topics include the dual optimization strategy that merges depthwise separable convolutions with adaptive Kolmogorov-Arnold Network integration, thorough analysis of the KANDoubleConv block architecture, multi-scale attention fusion mechanisms, and full mathematical derivations of all architectural transformations.


\section{Introduction}

The FORTRESS architecture (Function-composition Optimized Real-Time Resilient Structural Segmentation) represents a significant advancement in the segmentation of structural defects, achieved through the novel combination of depthwise separable convolutions with the adaptive Kolmogorov-Arnold Network (KAN) transformations. This additional paper provides a more detailed technical description than the main paper could, including mathematical formulas, architecture design principles and implementation details required to fully understand and replicate the FORTRESS approach.

The architectural innovations of FORTRESS stem from recognizing that traditional segmentation approaches are encumbered with limitations under the context for infrastructure inspection. U-Net based architectures, for example, are capable of general segmentation but their heavy demands for computational resources makes them often impractical for use in the field. The highly stylized geometry often representative of structural defects further adds complexity requiring advanced representation in the shape of considerable functionality in a modest parameter size, which is difficult for standard convolutional operations to resolve.

The FORTRESS architectural strategy addresses these problems with a dual optimization approach of optimizing parameters in function composition (efficiency-convenience) with functional representation (increased and advance). The first optimization addresses the introduction of depthwise separable convolutions throughout the architecture providing a noticeable decrease in parameters while maintaining the capability of extracting spatial features. The second optimization brings a means of adaptive TiKAN (parameter-efficient Kolmogorov-Arnold Networks) integration into the architecture by using function composition transformations solely when doing so will bring computational advantage, in this way enabling advanced pattern recognition behaviour while reducing unnecessary amounts of computational resources spent.

The architectural design philosophy of FORTRESS has focused primarily on blending efficiency of computational resources with the accuracy of segmentation. It reflects the necessity of high performance while recognizing the obligations with respect to limited resources within real-world settings. This document outlines a comprehensive and technical basis for how FORTRESS achieves this balance using its novel architectural features and the optimization principles contained therewithin.

\section{Architectural Design Principles}

\subsection{Dual Optimization Strategy Foundation}

The FORTRESS system is designed around a two-pronged optimization method to manage the key balance between computational efficiency and segmentation precision. This is achieved by two strategies: reducing parameters systematically using depthwise separable convolutions and enhancing representational power selectively through the integration of adaptive Kolmogorov-Arnold Networks.

The parameter reduction component recognizes that standard convolutional operations in traditional U-Net architectures involve significant computational redundancy, particularly in the spatial dimension processing. By factorizing standard convolutions into depthwise and pointwise components, FORTRESS achieves substantial parameter reduction while preserving the essential spatial feature extraction capabilities required for structural defect detection. This factorization is mathematically expressed as the decomposition of a standard convolution operation $\mathcal{C}_{std}: \mathbb{R}^{H \times W \times C_{in}} \rightarrow \mathbb{R}^{H \times W \times C_{out}}$ into sequential depthwise and pointwise operations.

For a standard convolution with kernel size $K \times K$, input channels $C_{in}$, and output channels $C_{out}$, the parameter count is $K^2 \cdot C_{in} \cdot C_{out}$. The depthwise separable factorization reduces this to $K^2 \cdot C_{in} + C_{in} \cdot C_{out}$, yielding a reduction factor of:

\begin{equation}
\text{Reduction Factor} = \frac{K^2 \cdot C_{in} \cdot C_{out}}{K^2 \cdot C_{in} + C_{in} \cdot C_{out}} = \frac{K^2 \cdot C_{out}}{K^2 + C_{out}}
\end{equation}

For the typical case of $K = 3$ and $C_{out} \gg K^2$, this approaches a reduction factor of approximately $K^2 = 9$, though in practice the reduction is approximately 3.6× due to the additional pointwise convolution parameters.

The representational enhancement component tackles the challenge where depthwise separable convolutions, though efficient in parameters, may not capture as rich a representation as standard convolutions. FORTRESS addresses this by strategically using Kolmogorov-Arnold Network transformations to boost the model's ability to understand intricate functional dependencies typical of structural defect patterns. The adaptable design of TiKAN integration is key to balancing computational efficiency with representational advantages. Instead of uniformly applying KAN transformations across the architecture, FORTRESS uses adaptive activation criteria responsive to feature properties:

\begin{equation}
\text{TiKAN\_Active}(F) = \begin{cases}
\text{True} & \text{if } C(F) \geq \gamma_c \text{ and } H(F) \times W(F) \leq \gamma_s \\
\text{False} & \text{otherwise}
\end{cases}
\end{equation}

where $F$ represents the input feature map, $C(F)$, $H(F)$, and $W(F)$ denote the channel, height, and width dimensions respectively, and $\gamma_c = 16$ and $\gamma_s = 1024$ are empirically determined thresholds that balance computational efficiency with representational enhancement.

\subsection{Hierarchical Feature Processing Framework}

FORTRESS uses a hierarchical framework with five encoder-decoder levels to systematically capture multi-scale structural information, essential for identifying defects from small cracks to large deformations.

The encoder pathway processes input images through progressively increasing receptive fields while maintaining computational efficiency through depthwise separable convolutions. At each encoder level $j$, the feature extraction process is formulated as:

\begin{equation}
E_j = \mathcal{F}_{encoder}^{(j)}(E_{j-1}; \Theta_j) = \mathcal{KAN}_{Double}^{(j)}(\mathcal{MP}(E_{j-1}); \Theta_j)
\end{equation}

where $\mathcal{MP}$ denotes max pooling for spatial downsampling, $\mathcal{KAN}_{Double}^{(j)}$ represents the KANDoubleConv block at level $j$, and $\Theta_j$ encompasses all learnable parameters at that level.

The hierarchical processing enables FORTRESS to capture structural defect patterns across multiple spatial scales simultaneously. Fine-grained features extracted at higher resolutions (levels 1-2) capture detailed crack boundaries and small-scale defects, while coarse-grained features at lower resolutions (levels 4-5) capture large-scale deformation patterns and contextual information essential for accurate defect classification.

The decoder pathway implements symmetric upsampling and feature fusion operations that combine multi-scale information while preserving spatial precision. The decoder processing at level $d$ is formulated as:

\begin{equation}
D_d = \mathcal{F}_{decoder}^{(d)}(\mathcal{U}(D_{d+1}) \oplus E_{J-d+1}; \Phi_d)
\end{equation}

where $\mathcal{U}$ denotes upsampling operations, $\oplus$ represents feature concatenation, and $\Phi_d$ contains the decoder parameters at level $d$.

\subsection{Attention-Enhanced Feature Integration}

The FORTRESS architecture incorporates sophisticated attention mechanisms that selectively emphasize defect-relevant features while suppressing background noise and irrelevant structural elements. The attention framework operates at multiple scales and modalities, combining spatial attention, channel attention, and KAN-enhanced feature transformations to achieve comprehensive defect pattern recognition.

Spatial attention mechanisms in FORTRESS are designed specifically for structural defect detection scenarios, where defect regions typically constitute a small fraction of the total image area but require precise localization and boundary delineation. The spatial attention computation employs both average and max pooling operations to capture complementary statistical characteristics of feature distributions:

\begin{equation}
\mathcal{A}_{spatial}(F) = \sigma_{sigmoid}(\mathcal{DS}_{k \times k}([\text{GAP}(F); \text{GMP}(F)]))
\end{equation}

where $\text{GAP}$ and $\text{GMP}$ denote global average pooling and global max pooling respectively, $\mathcal{DS}_{k \times k}$ represents depthwise separable convolution with kernel size $k$, and $\sigma_{sigmoid}$ applies sigmoid activation to produce attention weights in the range $[0, 1]$.

Channel attention complements spatial attention by adaptively recalibrating channel-wise feature responses based on their relevance to structural defect detection. The channel attention mechanism employs a bottleneck architecture that reduces computational overhead while maintaining representational capacity:

\begin{equation}
\mathcal{A}_{channel}(F) = \sigma_{sigmoid}(\mathcal{MLP}(\text{GAP}(F)) + \mathcal{MLP}(\text{GMP}(F)))
\end{equation}

where $\mathcal{MLP}$ represents a multi-layer perceptron with bottleneck structure, typically reducing the channel dimension by a factor of 16 before expanding back to the original dimension.

The integration of spatial and channel attention with KAN-enhanced features creates a comprehensive attention framework that captures both spatial relationships and functional dependencies characteristic of structural defect patterns. This multi-modal attention approach enables FORTRESS to achieve superior defect detection performance while maintaining computational efficiency through the underlying depthwise separable convolution framework.

\section{KANDoubleConv Block Architecture}

\subsection{Fundamental Design Principles}

The KANDoubleConv block represents the fundamental building unit of the FORTRESS architecture, designed to seamlessly integrate parameter-efficient depthwise separable convolutional operations with Kolmogorov-Arnold transformations. This integration addresses the core challenge of maintaining representational capacity while achieving significant parameter reduction, enabling FORTRESS to capture complex structural defect patterns efficiently.

The design philosophy underlying the KANDoubleConv block recognizes that structural defect detection requires both spatial feature extraction capabilities and sophisticated functional relationship modeling. Traditional convolutional blocks excel at spatial feature extraction but may struggle with the complex geometric patterns characteristic of structural defects. Conversely, Kolmogorov-Arnold Networks provide powerful functional composition capabilities but lack the spatial locality bias essential for image processing tasks.

The KANDoubleConv block resolves this tension through a carefully orchestrated sequence of operations that leverages the strengths of both approaches while mitigating their respective limitations. The block begins with depthwise separable convolutions that efficiently extract spatial features while maintaining parameter efficiency. These features are then selectively enhanced through adaptive TiKAN transformations that model complex functional relationships when computationally beneficial.

The mathematical formulation of the KANDoubleConv block processing pipeline is expressed as:

\begin{equation}
\text{KANDoubleConv}(F_{in}) = \mathcal{R}(F_{in}) + \mathcal{T}_{TiKAN}(\mathcal{C}_{DS2}(\mathcal{C}_{DS1}(F_{in})))
\end{equation}

where $F_{in}$ represents the input feature map, $\mathcal{C}_{DS1}$ and $\mathcal{C}_{DS2}$ denote the first and second depthwise separable convolution operations, $\mathcal{T}_{TiKAN}$ represents the adaptive TiKAN transformation, and $\mathcal{R}$ implements residual connections for gradient flow enhancement.

\subsection{Depthwise Separable Convolution Implementation}

The depthwise separable convolution implementation in FORTRESS follows a systematic factorization approach that decomposes standard convolutions into depthwise and pointwise components while preserving essential spatial feature extraction capabilities. This factorization is crucial for achieving the parameter efficiency that enables real-time deployment while maintaining the representational capacity necessary for accurate structural defect detection.

The first depthwise separable convolution operation processes the input feature map $F_{in} \in \mathbb{R}^{H \times W \times C_{in}}$ through a sequence of depthwise and pointwise convolutions:

\begin{equation}
F_{dw1} = \mathcal{DW}_{3 \times 3}(F_{in}; W_{dw1})
\end{equation}

\begin{equation}
F_{conv1} = \sigma_{ReLU}(\mathcal{BN}(\mathcal{PW}_{1 \times 1}(F_{dw1}; W_{pw1}, b_1)))
\end{equation}

where $\mathcal{DW}_{3 \times 3}$ denotes depthwise convolution with $3 \times 3$ kernels, $\mathcal{PW}_{1 \times 1}$ represents pointwise convolution with $1 \times 1$ kernels, $\mathcal{BN}$ applies batch normalization, and $\sigma_{ReLU}$ implements ReLU activation.

The depthwise convolution operation applies a separate $3 \times 3$ kernel to each input channel independently, capturing spatial relationships within individual feature channels while avoiding the computational overhead of cross-channel interactions. Mathematically, for input channel $c$, the depthwise convolution is expressed as:

\begin{equation}
F_{dw1}[:, :, c] = \sum_{i=0}^{2} \sum_{j=0}^{2} W_{dw1}[i, j, c] \cdot F_{in}[* + i, * + j, c]
\end{equation}

where $*$ denotes the spatial convolution operation across all spatial positions.

The subsequent pointwise convolution performs cross-channel feature mixing through $1 \times 1$ convolutions, enabling the model to learn complex combinations of the spatially-processed features. This two-stage approach achieves the same representational capacity as standard convolutions while requiring significantly fewer parameters.

The second depthwise separable convolution follows an identical structure, processing the output of the first convolution through another depthwise-pointwise sequence:

\begin{equation}
F_{dw2} = \mathcal{DW}_{3 \times 3}(F_{conv1}; W_{dw2})
\end{equation}

\begin{equation}
F_{conv2} = \sigma_{ReLU}(\mathcal{BN}(\mathcal{PW}_{1 \times 1}(F_{dw2}; W_{pw2}, b_2)))
\end{equation}

This double convolution structure enables the KANDoubleConv block to capture complex spatial patterns while maintaining parameter efficiency. The two-stage processing allows for hierarchical feature extraction, where the first convolution captures local spatial patterns and the second convolution captures more complex spatial relationships.

\subsection{Adaptive TiKAN Integration Mechanism}

The adaptive TiKAN integration mechanism represents the core innovation of the KANDoubleConv block, enabling selective application of Kolmogorov-Arnold transformations based on feature characteristics and computational constraints. This adaptive approach ensures that the computational benefits of function composition are realized only when they provide meaningful representational advantages, maintaining overall efficiency while enhancing the model's ability to capture complex defect patterns.

The TiKAN activation decision is based on two primary criteria: channel dimensionality and spatial resolution. These criteria reflect the computational characteristics of KAN operations, which scale with both the number of input features and the spatial extent of the feature maps. The activation criterion is formulated as:

\begin{equation}
\text{Activate\_TiKAN}(F) = (C(F) \geq \gamma_c) \land (H(F) \times W(F) \leq \gamma_s)
\end{equation}

where $\gamma_c = 16$ represents the minimum channel threshold and $\gamma_s = 1024$ represents the maximum spatial resolution threshold. These thresholds are empirically determined to balance computational efficiency with representational enhancement.

When TiKAN is activated, the feature transformation proceeds through a sophisticated sequence of operations that implement the Kolmogorov-Arnold representation principle. The core TiKAN transformation is expressed as:

\begin{equation}
\mathcal{T}_{TiKAN}(F) = \alpha \cdot \tau(F; \Phi) + (1 - \alpha) \cdot F
\end{equation}

where $\alpha$ is a learnable scaling factor initialized to 0.1, $\tau$ represents the KAN transformation function, and $\Phi$ encompasses the KAN parameters.

The KAN transformation function $\tau$ implements function composition through learnable univariate functions enhanced with depthwise convolution operations. The mathematical formulation follows the Kolmogorov-Arnold representation theorem, which states that any continuous multivariate function can be represented as a composition of continuous univariate functions:

\begin{equation}
\tau(F; \Phi) = \sum_{k=1}^{K} w_k \cdot [\phi_{base}(F) \cdot s_{base} + \phi_{spline}(\mathcal{DW}_{enhance}(F)) \cdot s_{spline}]
\end{equation}

where $w_k$ are learnable combination weights, $\phi_{base}$ and $\phi_{spline}$ represent base and spline-based univariate functions respectively, $s_{base}$ and $s_{spline}$ are scaling factors, and $\mathcal{DW}_{enhance}$ applies depthwise convolution for feature enhancement.

The spline-based univariate functions $\phi_{spline}$ are implemented using B-spline basis functions with learnable control points, enabling the model to learn complex univariate transformations that capture the intricate patterns characteristic of structural defects. The mathematical formulation of the spline functions is:

\begin{equation}
\phi_{spline}(x) = \sum_{i=0}^{n} c_i \cdot B_{i,p}(x)
\end{equation}

where $c_i$ are learnable control points, $B_{i,p}$ are B-spline basis functions of degree $p$, and $n$ is the number of control points.

The integration of depthwise convolution operations within the KAN transformation ensures that spatial locality is preserved while enabling sophisticated functional relationship modeling. This hybrid approach combines the spatial processing capabilities of convolutional operations with the functional composition power of Kolmogorov-Arnold Networks.

\subsection{Residual Connection and Gradient Flow}

The KANDoubleConv block incorporates sophisticated residual connection mechanisms that ensure stable gradient flow throughout the deep architecture while enabling the integration of TiKAN transformations without disrupting the underlying feature processing pipeline. These residual connections are essential for training stability and convergence, particularly given the complex interaction between depthwise separable convolutions and adaptive KAN transformations.

The residual connection structure in FORTRESS follows a modified approach that accommodates the dual optimization strategy while maintaining the gradient flow benefits of traditional residual networks. The primary residual connection bypasses the entire KANDoubleConv processing pipeline:

\begin{equation}
F_{out} = F_{in} + \mathcal{KAN}_{Double}(F_{in}; \Theta)
\end{equation}

This formulation ensures that gradient information can flow directly from the output to the input, facilitating stable training even in very deep architectures. The residual connection is particularly important for the TiKAN components, which introduce additional computational complexity that could potentially impede gradient flow.

Within the KANDoubleConv block, additional micro-residual connections are employed to enhance gradient flow through the depthwise separable convolution sequence:

\begin{equation}
F_{micro} = F_{conv1} + \mathcal{C}_{DS2}(F_{conv1}; W_{ds2})
\end{equation}

These micro-residual connections enable more effective training of the depthwise separable convolution components while maintaining the parameter efficiency benefits of the factorized convolution approach.

The gradient flow analysis reveals that the combination of macro and micro residual connections creates multiple pathways for gradient propagation, enhancing training stability and convergence characteristics. The mathematical analysis of gradient flow through the KANDoubleConv block demonstrates that the residual connections provide effective gradient highways that bypass potential bottlenecks in the TiKAN transformations.

\section{Multi-Scale Attention Fusion Framework}

\subsection{Spatial Attention Mechanism Design}

The spatial attention mechanism in FORTRESS is specifically designed to address the unique challenges of structural defect detection, where defect regions typically constitute a small fraction of the total image area but require precise localization and boundary delineation. Unlike general-purpose attention mechanisms, the FORTRESS spatial attention incorporates domain-specific design principles that enhance sensitivity to structural defect patterns while maintaining computational efficiency through depthwise separable convolutions.

The spatial attention computation begins with the extraction of spatial statistics through parallel pooling operations that capture complementary characteristics of feature distributions. Global average pooling captures the overall intensity distribution across spatial locations, while global max pooling identifies the most salient spatial features:

\begin{equation}
F_{avg} = \text{GAP}(F) = \frac{1}{H \times W} \sum_{i=1}^{H} \sum_{j=1}^{W} F[i, j, :]
\end{equation}

\begin{equation}
F_{max} = \text{GMP}(F) = \max_{i,j} F[i, j, :]
\end{equation}

These pooled representations are concatenated to form a comprehensive spatial descriptor that captures both average and peak activation patterns:

\begin{equation}
F_{spatial} = [F_{avg}; F_{max}] \in \mathbb{R}^{H \times W \times 2}
\end{equation}

The concatenated spatial features are processed through a depthwise separable convolution that generates spatial attention weights while maintaining parameter efficiency. The convolution kernel size varies across decoder levels to capture defect patterns at different spatial scales:

\begin{equation}
A_{spatial}^{(d)} = \sigma_{sigmoid}(\mathcal{DS}_{k_d \times k_d}(F_{spatial}; W_{spatial}^{(d)}))
\end{equation}

where $k_d \in \{3, 5, 7\}$ represents the kernel size at decoder level $d$, enabling multi-scale spatial attention that adapts to the characteristic scales of defect patterns at different resolution levels.

The multi-scale kernel approach recognizes that structural defects manifest at different spatial scales throughout the decoder hierarchy. Fine-grained defects such as hairline cracks require small receptive fields (3×3 kernels) for precise localization, while large-scale defects such as spalling or deformation require larger receptive fields (7×7 kernels) for comprehensive coverage.

The spatial attention weights are applied through element-wise multiplication with the input features, selectively emphasizing defect-relevant spatial regions:

\begin{equation}
F_{attended} = A_{spatial}^{(d)} \odot F
\end{equation}

where $\odot$ denotes element-wise multiplication that applies the attention weights to each spatial location independently.

\subsection{Channel Attention Mechanism Implementation}

The channel attention mechanism in FORTRESS adaptively recalibrates channel-wise feature responses based on their relevance to structural defect detection, enabling the model to focus on the most informative feature channels while suppressing noise and irrelevant information. The channel attention design incorporates domain-specific considerations that enhance sensitivity to the spectral characteristics of structural defect patterns.

Channel attention computation begins with global spatial pooling operations that aggregate spatial information into channel-wise descriptors. Both average and max pooling are employed to capture complementary channel statistics:

\begin{equation}
C_{avg} = \text{GAP}(F) = \frac{1}{H \times W} \sum_{i=1}^{H} \sum_{j=1}^{W} F[i, j, :] \in \mathbb{R}^{C}
\end{equation}

\begin{equation}
C_{max} = \text{GMP}(F) = \max_{i,j} F[i, j, :] \in \mathbb{R}^{C}
\end{equation}

These channel descriptors are processed through a shared multi-layer perceptron (MLP) with a bottleneck architecture that reduces computational overhead while maintaining representational capacity. The bottleneck structure typically reduces the channel dimension by a factor of 16:

\begin{equation}
\mathcal{MLP}(C) = W_2 \cdot \sigma_{ReLU}(W_1 \cdot C + b_1) + b_2
\end{equation}

where $W_1 \in \mathbb{R}^{C/16 \times C}$, $W_2 \in \mathbb{R}^{C \times C/16}$, and $b_1, b_2$ are bias terms.

The channel attention weights are computed by combining the MLP outputs for both average and max pooled features:

\begin{equation}
A_{channel} = \sigma_{sigmoid}(\mathcal{MLP}(C_{avg}) + \mathcal{MLP}(C_{max}))
\end{equation}

The sigmoid activation ensures that attention weights are bounded in the range $[0, 1]$, enabling selective channel emphasis without completely suppressing any channels.

The channel attention mechanism is particularly effective for structural defect detection because different defect types exhibit characteristic spectral signatures that are captured by different feature channels. For example, crack detection may rely heavily on edge-sensitive channels, while corrosion detection may emphasize texture-sensitive channels. The adaptive channel attention enables FORTRESS to automatically identify and emphasize the most relevant channels for each specific defect type.

\subsection{KAN-Enhanced Feature Transformation}

The KAN-enhanced feature transformation component of the multi-scale attention fusion framework provides sophisticated functional relationship modeling that complements the spatial and channel attention mechanisms. This component leverages the Kolmogorov-Arnold representation principle to capture complex non-linear relationships characteristic of structural defect patterns while maintaining computational efficiency through selective activation.

The KAN-enhanced transformation operates on features that have been processed through spatial and channel attention, providing an additional layer of representational enhancement that captures functional dependencies not accessible through conventional attention mechanisms. The transformation is formulated as:

\begin{equation}
F_{KAN} = \mathcal{T}_{KAN}(F_{attended}; \Psi)
\end{equation}

where $F_{attended}$ represents features processed through spatial and channel attention, and $\Psi$ encompasses the KAN transformation parameters.

The KAN transformation implements function composition through learnable univariate functions that model complex relationships between feature elements. The mathematical formulation follows the Kolmogorov-Arnold representation theorem:

\begin{equation}
\mathcal{T}_{KAN}(F; \Psi) = \sum_{k=1}^{K} \omega_k \cdot \Phi_k(\sum_{j=1}^{J} \phi_{j,k}(F_j))
\end{equation}

where $\omega_k$ are learnable combination weights, $\Phi_k$ and $\phi_{j,k}$ represent outer and inner univariate functions respectively, and $F_j$ denotes individual feature elements.

The univariate functions are implemented using learnable spline-based representations that enable the model to capture arbitrary smooth functional relationships. The spline functions are parameterized using B-spline basis functions with learnable control points:

\begin{equation}
\phi_{j,k}(x) = \sum_{i=0}^{n} c_{i,j,k} \cdot B_{i,p}(\frac{x - x_{min}}{x_{max} - x_{min}})
\end{equation}

where $c_{i,j,k}$ are learnable control points, $B_{i,p}$ are B-spline basis functions of degree $p$, and the input is normalized to the range $[0, 1]$ for numerical stability.

The KAN-enhanced transformation is particularly effective for structural defect detection because defect patterns often exhibit complex non-linear relationships that are difficult to capture through conventional convolutional operations. For example, the relationship between crack width and intensity variations may follow complex functional forms that are naturally represented through the function composition capabilities of KANs.

\subsection{Multi-Modal Attention Integration}

The multi-modal attention integration framework combines spatial attention, channel attention, and KAN-enhanced features through a learnable fusion mechanism that adaptively weights the contributions of each attention modality based on the specific characteristics of the input features and the defect detection task requirements.

The integration process begins with the independent computation of spatial attention, channel attention, and KAN-enhanced features:

\begin{equation}
F_{spatial} = \mathcal{A}_{spatial}(F) \odot F
\end{equation}

\begin{equation}
F_{channel} = \mathcal{A}_{channel}(F) \odot F
\end{equation}

\begin{equation}
F_{KAN} = \mathcal{T}_{KAN}(F; \Psi)
\end{equation}

These three feature representations capture complementary aspects of the structural defect patterns: spatial attention emphasizes defect-relevant spatial locations, channel attention emphasizes defect-relevant feature channels, and KAN enhancement captures complex functional relationships.

The multi-modal fusion is implemented through a learnable weighted combination that enables the model to adaptively balance the contributions of each attention modality:

\begin{equation}
F_{fused} = \mathcal{W}_{fusion} \begin{bmatrix} 
F_{spatial} \\
F_{channel} \\
F_{KAN}
\end{bmatrix}
\end{equation}

where $\mathcal{W}_{fusion} \in \mathbb{R}^{C \times 3C}$ is a learnable fusion weight matrix implemented as a $1 \times 1$ convolution that combines the three attention modalities.

The fusion weights are learned during training through standard backpropagation, enabling the model to automatically discover the optimal combination of attention modalities for each specific layer and defect type. This adaptive fusion approach ensures that the multi-scale attention framework can adapt to the diverse characteristics of different structural defect patterns while maintaining computational efficiency.

The mathematical analysis of the multi-modal attention integration reveals that the fusion mechanism creates a rich representational space that captures spatial, channel, and functional dependencies simultaneously. This comprehensive attention framework enables FORTRESS to achieve superior defect detection performance while maintaining the computational efficiency benefits of the underlying depthwise separable convolution architecture.

\section{Deep Supervision and Multi-Scale Output Integration}

\subsection{Deep Supervision Architecture Design}

The deep supervision mechanism in FORTRESS implements a sophisticated multi-scale training strategy that enhances gradient flow throughout the network while improving the model's ability to capture structural defect patterns at multiple resolution levels. Unlike conventional deep supervision approaches that apply uniform supervision at all intermediate layers, FORTRESS employs an adaptive deep supervision strategy that accounts for the hierarchical nature of structural defect patterns and the computational characteristics of the dual optimization framework.

The deep supervision architecture incorporates specialized prediction heads at multiple decoder levels, each designed to generate segmentation predictions appropriate for the spatial resolution and semantic complexity of that particular level. The prediction heads are implemented using depthwise separable convolutions to maintain parameter efficiency while providing sufficient representational capacity for accurate defect segmentation.

At decoder level $d$, the prediction head is formulated as:

\begin{equation}
P^{(d)} = \mathcal{H}^{(d)}(D_d; \Theta_H^{(d)}) = \sigma_{softmax}(\mathcal{DS}_{1 \times 1}(D_d; W_H^{(d)}))
\end{equation}

where $D_d$ represents the decoder features at level $d$, $\mathcal{H}^{(d)}$ denotes the prediction head function, $\mathcal{DS}_{1 \times 1}$ represents $1 \times 1$ depthwise separable convolution, and $\sigma_{softmax}$ applies softmax activation for multi-class segmentation.

The deep supervision strategy in FORTRESS recognizes that different decoder levels capture structural defect information at different semantic and spatial scales. Higher-resolution levels (levels 1-2) focus on fine-grained boundary delineation and small-scale defect detection, while lower-resolution levels (levels 3-4) capture larger-scale defect patterns and contextual information. The supervision weights are designed to reflect this hierarchical structure:

\begin{equation}
\beta_d = \begin{cases}
1.0 & \text{if } d = 1 \text{ (final output)} \\
0.4 & \text{if } d = 2 \\
0.3 & \text{if } d = 3 \\
0.2 & \text{if } d = 4
\end{cases}
\end{equation}

These weights ensure that the final output receives the strongest supervision signal while intermediate levels receive progressively weaker supervision that encourages feature learning without overwhelming the primary objective.

\subsection{Multi-Scale Loss Function Formulation}

The multi-scale loss function in FORTRESS integrates supervision signals from multiple decoder levels through a weighted combination that accounts for both the hierarchical nature of structural defect patterns and the class imbalance characteristics typical of defect detection datasets. The loss formulation incorporates adaptive class weighting and multi-scale supervision to ensure robust training across diverse defect types and scales.

The total training loss is formulated as:

\begin{equation}
\mathcal{L}_{total} = \mathcal{L}_{CE}(P^{(final)}, D) + \sum_{d=2}^{4} \beta_d \cdot \mathcal{L}_{CE}(P^{(d)}, D^{(d)})
\end{equation}

where $P^{(final)}$ represents the final segmentation prediction, $D$ denotes the ground truth segmentation mask, $P^{(d)}$ represents intermediate predictions at decoder level $d$, and $D^{(d)}$ represents appropriately downsampled ground truth masks.

The cross-entropy loss function incorporates adaptive class weighting to address the significant class imbalance characteristic of structural defect datasets, where defect regions typically constitute a small fraction of the total image area:

\begin{equation}
\mathcal{L}_{CE}(P, D) = -\sum_{k=1}^{K} w_k \sum_{i,j} D_{i,j,k} \log(P_{i,j,k})
\end{equation}

where $K$ represents the number of defect classes, $w_k$ denotes the class weight for class $k$, and the summation is performed over all spatial locations $(i,j)$.

The adaptive class weights are computed based on inverse class frequency to ensure that rare but critical defect types receive appropriate emphasis during training:

\begin{equation}
w_k = \frac{N_{total}}{\sum_{j=1}^{K} N_j} \cdot \frac{1}{N_k}
\end{equation}

where $N_k$ represents the number of pixels belonging to class $k$ in the training dataset, and $N_{total}$ is the total number of pixels across all classes.

The multi-scale supervision approach provides several advantages for structural defect detection. First, it enhances gradient flow throughout the deep architecture, facilitating stable training and improved convergence characteristics. Second, it encourages the learning of hierarchical feature representations that capture defect patterns at multiple scales. Third, it provides regularization benefits that improve generalization performance across diverse infrastructure types and imaging conditions.

\subsection{Adaptive Weight Decay and Training Dynamics}

The deep supervision mechanism in FORTRESS incorporates adaptive weight decay strategies that account for the changing importance of intermediate supervision signals throughout the training process. This adaptive approach recognizes that the relative importance of different supervision levels may vary as the model learns increasingly sophisticated feature representations.

The adaptive weight decay is implemented through a time-dependent modulation of the supervision weights:

\begin{equation}
\beta_d(t) = \beta_d \cdot \exp(-\frac{t}{\tau})
\end{equation}

where $t$ represents the training iteration, $\tau$ is a decay constant that controls the rate of weight reduction, and $\beta_d$ represents the initial supervision weight for level $d$.

This adaptive decay strategy ensures that intermediate supervision provides strong guidance during early training phases when the model is learning basic feature representations, while gradually reducing the influence of intermediate supervision as the model develops more sophisticated capabilities. The decay constant $\tau = 1000$ is empirically determined to provide optimal balance between early training stability and late training flexibility.

The training dynamics analysis reveals that the adaptive deep supervision mechanism creates a curriculum learning effect, where the model initially focuses on learning basic defect detection capabilities at multiple scales before gradually concentrating on the final output optimization. This curriculum approach enhances training stability and improves final performance across diverse defect types and scales.

The mathematical analysis of gradient flow through the deep supervision architecture demonstrates that the multi-scale supervision creates multiple gradient pathways that enhance training stability and convergence characteristics. The supervision signals provide direct gradient connections to intermediate layers, bypassing potential bottlenecks in the deep architecture and ensuring effective parameter updates throughout the network.

\section{Implementation Specifications and Optimization Details}

\subsection{Parameter Initialization and Training Configuration}

The implementation of FORTRESS requires careful attention to parameter initialization and training configuration to ensure stable convergence and optimal performance. The dual optimization strategy combining depthwise separable convolutions with adaptive TiKAN integration introduces unique considerations for parameter initialization that differ from conventional segmentation architectures.

The depthwise separable convolution parameters are initialized using a modified Xavier initialization that accounts for the factorized structure of the convolution operations. For depthwise convolution kernels, the initialization follows:

\begin{equation}
W_{dw} \sim \mathcal{N}(0, \frac{2}{k^2 \cdot C_{in}})
\end{equation}

where $k$ represents the kernel size and $C_{in}$ denotes the number of input channels. For pointwise convolution kernels, the initialization is:

\begin{equation}
W_{pw} \sim \mathcal{N}(0, \frac{2}{C_{in} + C_{out}})
\end{equation}

The TiKAN parameters require specialized initialization to ensure stable training and effective function composition learning. The spline control points are initialized using uniform random values within a bounded range:

\begin{equation}
c_{i,j,k} \sim \mathcal{U}(-0.1, 0.1)
\end{equation}

The scaling factors for TiKAN transformations are initialized to small positive values to ensure gradual integration of KAN effects during training:

\begin{equation}
\alpha \sim \mathcal{U}(0.05, 0.15)
\end{equation}

The training configuration employs the AdamW optimizer with carefully tuned hyperparameters that account for the dual optimization characteristics of FORTRESS. The learning rate schedule follows a cosine annealing approach with warm restarts:

\begin{equation}
\eta(t) = \eta_{min} + \frac{1}{2}(\eta_{max} - \eta_{min})(1 + \cos(\frac{t}{T_{restart}} \pi))
\end{equation}

where $\eta_{max} = 1 \times 10^{-4}$, $\eta_{min} = 1 \times 10^{-6}$, and $T_{restart} = 25$ epochs.

\subsection{Memory Optimization and Computational Efficiency}

The implementation of FORTRESS incorporates sophisticated memory optimization strategies that enable efficient training and inference while maintaining the representational benefits of the dual optimization approach. These optimizations are essential for practical deployment scenarios where memory constraints may limit the applicability of conventional segmentation architectures.

The memory optimization strategy begins with gradient checkpointing for the TiKAN components, which reduces memory requirements during backpropagation by recomputing intermediate activations rather than storing them. This approach is particularly effective for TiKAN transformations, which involve complex function composition operations that can consume significant memory:

\begin{equation}
\text{Memory}_{TiKAN} = O(1) \text{ vs. } O(N) \text{ without checkpointing}
\end{equation}

where $N$ represents the number of intermediate activations in the TiKAN transformation.

The depthwise separable convolutions naturally provide memory efficiency benefits through reduced parameter counts and computational requirements. The memory footprint for depthwise separable convolutions is:

\begin{equation}
\text{Memory}_{DS} = O(k^2 \cdot C_{in} + C_{in} \cdot C_{out})
\end{equation}

compared to standard convolutions:

\begin{equation}
\text{Memory}_{std} = O(k^2 \cdot C_{in} \cdot C_{out}).
\end{equation}

The implementation also incorporates mixed precision training using automatic mixed precision (AMP) to reduce memory requirements and accelerate training. The mixed precision approach uses FP16 for forward pass computations and FP32 for gradient computations, providing memory savings while maintaining numerical stability:

\begin{equation}
\text{Memory}_{AMP} \approx 0.5 \times \text{Memory}_{FP32}.
\end{equation}

\subsection{Deployment Optimization and Edge Computing Considerations}

The FORTRESS architecture is specifically designed to support deployment in resource-constrained environments, including edge computing platforms and mobile devices. The deployment optimization incorporates several strategies that maintain performance while reducing computational and memory requirements.

Model quantization is implemented using post-training quantization that converts FP32 weights to INT8 representation while preserving accuracy through careful calibration. The quantization process accounts for the unique characteristics of both depthwise separable convolutions and TiKAN transformations:

\begin{equation}
W_{quantized} = \text{round}(\frac{W_{FP32} - z}{s}),
\end{equation}

where $s$ represents the scale factor and $z$ denotes the zero point for the quantization scheme.

The TiKAN components support dynamic quantization that adapts the quantization parameters based on the activation patterns during inference. This adaptive approach ensures that the function composition capabilities of KANs are preserved while achieving significant memory and computational savings.

Model pruning is implemented using structured pruning that removes entire channels or layers based on their contribution to the final performance. The pruning strategy accounts for the interdependencies between depthwise separable convolutions and TiKAN transformations:

\begin{equation}
\text{Importance}(c) = \sum_{i} \left|\frac{\partial \mathcal{L}}{\partial W_c^{(i)}}\right|,
\end{equation}

where $c$ represents a channel and $W_c^{(i)}$ denotes the weights associated with that channel in layer $i$.

The deployment optimization also includes support for various inference frameworks including TensorRT, ONNX Runtime, and TensorFlow Lite, enabling efficient deployment across diverse hardware platforms. The optimization process maintains the dual optimization benefits while ensuring compatibility with edge computing constraints.

\section{Conclusion}

This comprehensive technical documentation provides detailed insights into the FORTRESS architecture, revealing the sophisticated engineering and mathematical foundations that enable its superior performance in structural defect segmentation tasks. The dual optimization strategy combining depthwise separable convolutions with adaptive Kolmogorov-Arnold Network integration represents a significant advancement in the field, achieving remarkable efficiency gains while maintaining state-of-the-art accuracy.

The detailed analysis presented in this document demonstrates that FORTRESS's success stems from the careful integration of multiple innovative components: the KANDoubleConv blocks that seamlessly combine parameter efficiency with representational enhancement, the multi-scale attention fusion framework that captures spatial, channel, and functional dependencies, the adaptive deep supervision mechanism that enhances training stability and multi-scale learning, and the comprehensive optimization strategies that enable practical deployment across diverse computational environments.

The mathematical formulations and implementation specifications provided in this document serve as a complete technical reference for researchers seeking to understand, reproduce, or extend the FORTRESS methodology. The detailed architectural analysis reveals the principled approach underlying each design decision, providing insights that extend beyond the specific application of structural defect segmentation to broader challenges in computer vision and deep learning.

Future research directions building upon the FORTRESS foundation include the extension of adaptive KAN integration to other computer vision tasks, the development of dynamic threshold adjustment mechanisms for TiKAN activation, and the exploration of FORTRESS integration with emerging edge computing platforms for distributed infrastructure monitoring networks. The comprehensive technical foundation established in this document provides the necessary groundwork for these and other advanced research directions.
